\documentclass{article}

\usepackage{algorithm}
\usepackage{algorithmic}
\usepackage{amsmath}
\usepackage{booktabs}
\usepackage{float}
\usepackage{geometry}
\usepackage{graphicx}
\usepackage{hyperref}
\usepackage{listings}
\usepackage[lighttt]{lmodern}
\usepackage{maths}
\usepackage{microtype}
\usepackage{subfigure}
\usepackage{url}
\usepackage{xcolor}

\hypersetup{
    colorlinks,
    linkcolor={red!50!black},
    urlcolor={blue!80!black}
}

\DeclareMathOperator{\argmax}{argmax}

\newcommand{\emaillink}[1]{\href{#1}{\nolinkurl{#1}}}

\newcommand{\imagepath}[0]{images}

\newcommand{\stufig}[5]
{
        \begin{figure}[#5]
        \begin{center}
                \includegraphics[#1]{#2}
                \caption{#3}
                \label{#4}
        \end{center}
        \end{figure}
}

\floatstyle{ruled}
\newfloat{stulisting}{thp}{lop}
\floatname{stulisting}{Listing}

\lstdefinestyle{Default}{
abovecaptionskip=0.5cm,
basicstyle=\scriptsize\ttfamily,
belowcaptionskip=0.5cm,
belowskip=0.5cm,
columns=fixed,
commentstyle=\slshape, 
keywordstyle=\bfseries,
language=C++,
numbers=none, 
numbersep=5pt,
numberstyle=\tiny,
mathescape=true,
showstringspaces=false,
stepnumber=1,
tabsize=2
}

\begin{document}

\title{SemanticPaint: A Framework for the \\ Interactive Segmentation of 3D Scenes}
\author{
Stuart Golodetz$^*$ \\ \emaillink{sgolodetz@gxstudios.net} \\ University of Oxford, UK
\and
Michael Sapienza$^*$ \\ \emaillink{michael.sapienza@eng.ox.ac.uk} \\ University of Oxford, UK
\and
Julien P. C. Valentin \\ \emaillink{julien.valentin@eng.ox.ac.uk} \\ University of Oxford, UK
\and
Vibhav Vineet \\ \emaillink{vibhav.vineet@gmail.com} \\ Stanford University, USA
\and
Ming-Ming Cheng \\ \emaillink{cmm.thu@qq.com} \\ Nankai University, China
\and
Anurag Arnab \\ \emaillink{anurag.arnab@gmail.com} \\ University of Oxford, UK
\and
Victor A. Prisacariu \\ \emaillink{victor@viprad.net} \\ University of Oxford, UK
\and
Olaf K{\"a}hler \\ \emaillink{olaf@robots.ox.ac.uk} \\ University of Oxford, UK
\and
Carl Yuheng Ren \\ \emaillink{ren@carlyuheng.com} \\ University of Oxford, UK
\and
David W. Murray \\ \emaillink{david.murray@eng.ox.ac.uk} \\ University of Oxford, UK
\and
Shahram Izadi \\ \emaillink{shahrami@microsoft.com} \\ Microsoft Research, USA
\and
Philip H. S. Torr \\ \emaillink{philip.torr@eng.ox.ac.uk} \\ University of Oxford, UK
}
\date{\today}
\maketitle

\begin{center}
\includegraphics[width=.9\linewidth]{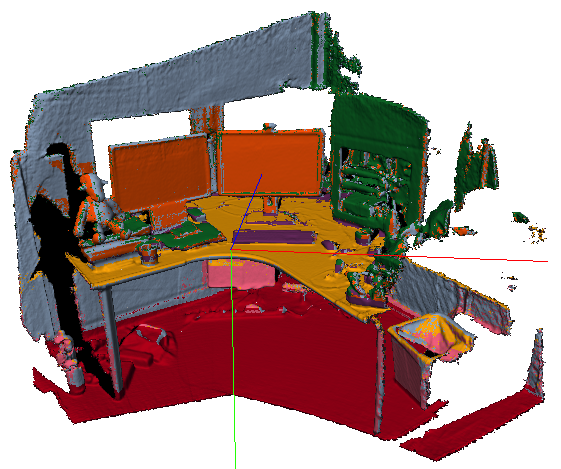}
\end{center}

\newpage

\tableofcontents

\newpage

\section{Introduction}

Scene understanding, the overarching goal of allowing a computer to understand what it sees in an image and thereby derive some kind of holistic meaning about the scene being viewed, is a fundamental problem that spans the entire field of computer vision.
As such, it has been the subject of intense research ever since computer vision was na{\"i}vely assigned to Gerald Sussman as a `summer project' in 1966.
However, it remains a notoriously difficult problem even to specify, let alone solve in a general sense.

What does it mean for a computer to \emph{understand} a scene?
We might think it desirable for a computer to be able to take in a scene at a glance and infer not only what people and objects are present in the scene and where they are, but what they are doing and their relationships to each other.
We might even hope to be able to predict what might happen next.
In reality, even the first of these (the problem of determining what is present in the scene and where it is) is far from being solved.
In spite of the progress that has been made in recent years, fully understanding an arbitrary scene in an automatic way remains a major challenge.
Similarly, although many approaches have taken into account the relationships between objects in a scene (e.g.~\cite{Choi2015,Lin2013,Zheng2015}), we remain far from the overall goal of fully understanding the interactions between them and making predictions about what might be about to happen.

These tasks remain fundamental challenges for the computer vision community as a whole, and we do not pretend to be able to solve them here.
We mention them primarily to illustrate the extent to which scene understanding is likely to remain a significant research challenge over the coming years, and to highlight the continuing need for foundational work in this area that other researchers can build upon to help drive the field forwards.

In this report, we present our contribution to this foundational work in the form of an interactive segmentation framework for 3D scenes that we call \emph{SemanticPaint}~\cite{Valentin2015,Golodetz2015}.
Using our framework, a user can walk into a room wearing a depth camera and a virtual reality headset, and both densely reconstruct the 3D scene~\cite{Niessner2013,Kaehler2015} and interactively segment the environment into object classes such as `chair', `floor' and `table'.
The user interacts \emph{physically} with the real-world scene, touching objects and using voice commands to assign them appropriate labels.
These user-generated labels are leveraged by an online random forest-based machine learning algorithm, which is used to predict labels for previously unseen parts of the scene.
The entire pipeline runs in real time, and the user stays `in the loop' throughout the process, receiving immediate feedback about the progress of the labelling and interacting with the scene as necessary to refine the predicted segmentation.

Since we keep the user in the loop, our framework can be used to produce high-quality, \emph{personalised} segmentations of a real-world environment.
Such segmentations have numerous uses, e.g.~(i) we can use them to identify walkable surfaces in an environment as part of the process of generating a navigation map that can provide routing support to people or robots; (ii) we can use them to help partially-sighted people avoid collisions by highlighting the obstacles in an environment; and (iii) in a computer vision setting, we can extract 3D models from them that can be used to train object detectors.

By releasing our work in library form, we hope to make it straightforward for other people to implement ideas such as this on top of our existing work, rather than having to implement the scene labelling pipeline for each new application from scratch.

\section{Architectural Overview}

\stufig{height=7cm}{\imagepath/highlevelarchitecture}{The high-level architecture of the \emph{SemanticPaint} framework. The framework is organised as a set of reusable libraries (\texttt{rafl}, \texttt{rigging}, \texttt{spaint} and \texttt{tvgutil}), together with an application (\texttt{spaintgui}) that illustrates how to make use of these libraries to implement an interactive 3D scene segmentation tool.}{fig:highlevelarchitecture}{!t}

The high-level architecture of the \emph{SemanticPaint} framework is shown in Figure~\ref{fig:highlevelarchitecture}.
It is organised as a set of reusable libraries (\texttt{rafl}, \texttt{rigging}, \texttt{spaint} and \texttt{tvgutil}), together with an application (\texttt{spaintgui}) that illustrates how to make use of these libraries to implement an interactive 3D scene segmentation tool.
In terms of functionality, \texttt{rafl} contains our generic implementation of random forests, \texttt{rigging} contains our implementation of camera rigs, \texttt{spaint} contains the main \emph{SemanticPaint} functionality (sampling, feature calculation, label propagation and user interaction) and \texttt{tvgutil} contains support code.

In this section, we will first take a top-down view of the \texttt{spaintgui} application in order to make clear the way in which the framework is organised, and then take a closer look at the structures of some of the more important libraries.
We defer a detailed discussion of the various different parts of the system until later in the report.

\subsection{The \texttt{spaintgui} Application}
\label{subsec:spaintgui}

As previously mentioned, \texttt{spaintgui} is a tool for interactive 3D scene segmentation.
It is structured in much the same way as a conventional video game, i.e.~it performs an infinite loop, each iteration of which processes user input, updates the scene and renders a frame (see Figure~\ref{fig:applicationloop}).
The application has a number of different modes (see Table~\ref{tbl:modes}) that the user will manually switch between in order to effectively reconstruct and label a scene.
A typical \texttt{spaintgui} workflow might involve:
\begin{enumerate}
\item \textbf{Reconstruction}. Starting in normal mode, the user reconstructs a 3D model of the scene. Reconstruction in \emph{SemanticPaint} is performed using the InfiniTAM v2 fusion engine of K{\"a}hler \emph{et al.} \cite{Kaehler2015}.
\item \textbf{Labelling}. The user chooses a method for labelling the scene (see \S\ref{subsec:selectors}) and switches to propagation mode. Using their chosen method, they provide ground truth labels at sparse locations in the scene, which are then propagated over smooth surfaces to augment the training data the user is providing (see \S\ref{subsec:labelpropagation}). For example, the user might choose to interact with the scene using touch, in which case providing labels would involve physically touching objects in the scene.
\item \textbf{Training/Prediction}. Having labelled relevant parts of the scene, the user now switches to training mode to learn a model (a random forest) of the object classes present in the scene (see \S\ref{subsec:training}). Once a model has been learnt, the user switches to prediction mode to predict labels for currently-unlabelled parts of the scene (see \S\ref{subsec:prediction}).
\item \textbf{Correction}. Finally, the user switches to training-and-prediction mode, and corrects the labelling of the scene as necessary using their chosen input method.
\end{enumerate}
An illustration of the first three steps of this process is shown in Figure~\ref{fig:workflow}. For an example of correction, please see the videos on our \url{http://www.semantic-paint.com} project page.

\subsubsection{Processing User Input}

The first step in each iteration of the application's main loop is to process any user input provided, so as to allow the user to move the camera around, label the scene, undo/redo previous labellings, or change the application mode, renderer or current label. The details of how we process the input are largely uninteresting, so we will not dwell on them, but readers who are interested in our implementations of camera control or command management may want to look at Appendices~\ref{sec:cameracontrol} and \ref{sec:commandmanagement}. It should also be noted that parts of the application's functionality (e.g.~changing the application mode or current label) can be controlled via voice commands: details of how voice recognition is implemented can be found in Appendix~\ref{sec:voicerecognition}.

\stufig{height=7cm}{\imagepath/applicationloop}{The main application loop of \texttt{spaintgui}, which is structured in much the same way as a conventional video game.}{fig:applicationloop}{!t}

\begin{table}[!t]
\centering
\footnotesize
\begin{tabular}{ll}
\toprule
\textbf{Mode} & \textbf{Purpose} \\
\midrule
Normal & Allows the user to reconstruct a 3D model of the scene and label voxels \\
Propagation & Normal + propagates the current label across smooth surfaces in the scene \\
Training & Normal + trains a random forest to predict labels for unlabelled scene voxels \\
Prediction & Normal + uses the forest to predict labels for unlabelled scene voxels \\
Training-and-Prediction & Normal + interleaves training and prediction on alternate frames \\
Feature Inspection & Normal + allows the user to inspect voxel features (requires OpenCV) \\
\bottomrule
\end{tabular}
\normalsize
\caption{The different modes supported in the \texttt{spaintgui} application.}
\label{tbl:modes}
\end{table}

\stufig{width=\linewidth}{\imagepath/workflow}{The first few steps of a typical \texttt{spaintgui} workflow (see \S\ref{subsec:spaintgui}).}{fig:workflow}{!t}

\subsubsection{Updating the Scene}
\label{subsubsec:updatescene}

\stufig{height=15cm}{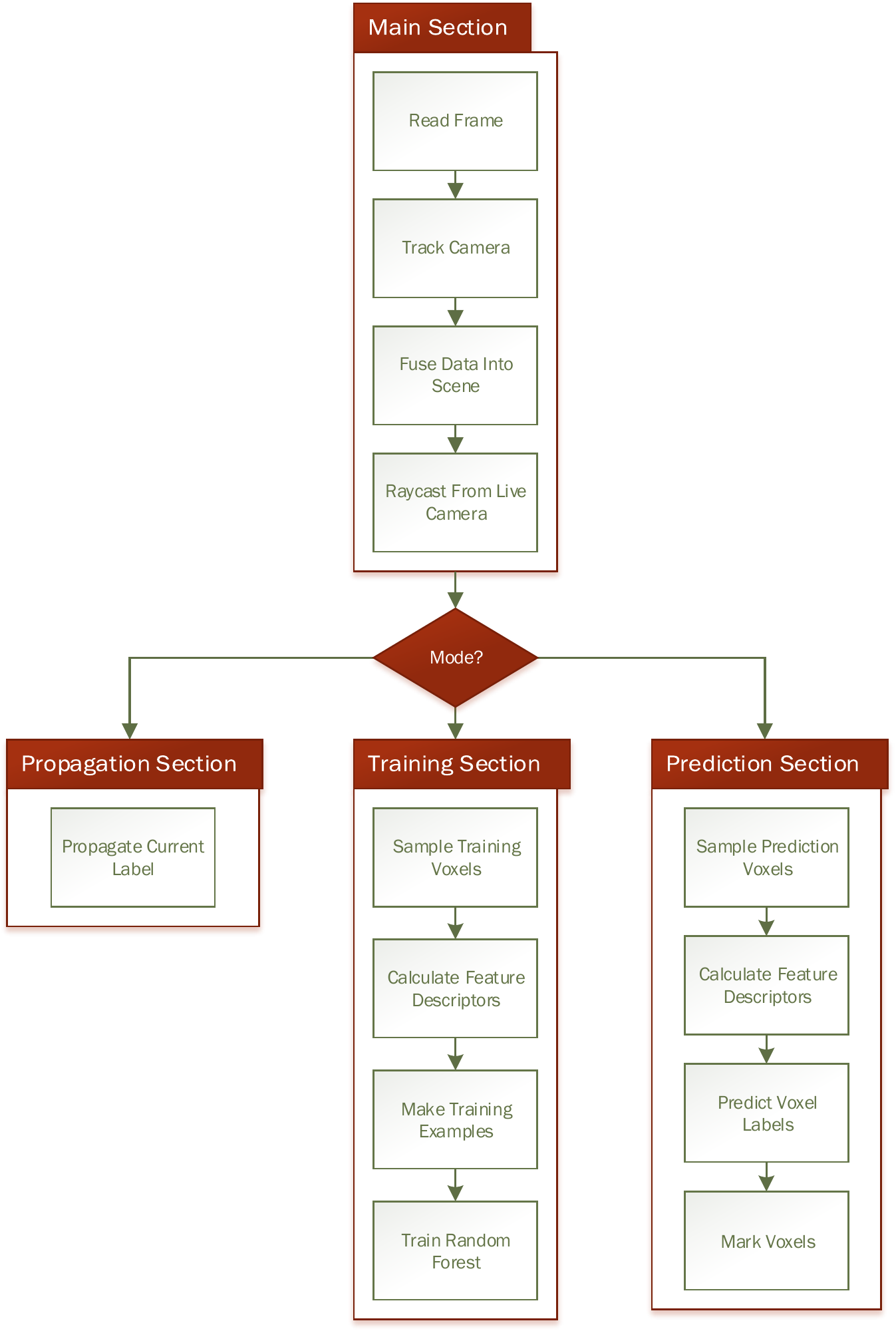}{The processing pipeline used to update the scene.}{fig:pipelinearchitecture}{!t}

Our mechanism for updating the scene is structured as a processing pipeline, which we divide into a number of different sections (see Figure~\ref{fig:pipelinearchitecture}).
In each frame, only some of the sections of the pipeline will be run, depending on the current mode of the system.
In most modes, the main section and a section of the pipeline corresponding to the mode in question will be run in sequence.
However, in \emph{training-and-prediction} mode, we interleave training with prediction by running the corresponding sections on alternate frames: this allows us to achieve the effect of performing training and prediction `simultaneously' without having a detrimental impact on the frame rate of the overall system.

\paragraph{Main Section}

The main section of the pipeline performs geometric scene reconstruction and camera pose estimation, and is run on every frame (although it is possible to stop reconstructing the scene and only estimate the pose of the camera in the currently-reconstructed scene if desired).
The first step in each frame is to read new depth and colour images from the current image source (generally either a depth camera such as the Asus Xtion or Kinect, or a directory on disk).
These are then passed to whichever tracker is in use to allow it to determine an estimate of the camera's pose in the new frame.
If reconstruction is enabled, the next step is then to update the global scene using the information from the depth and colour images (bearing in mind that their location in relation to the scene is now known): this process is generally known as \emph{fusion}.
Finally, a raycast is performed from the current estimated camera pose: this will be needed as an input to the tracker in the next frame.
For more details about how InfiniTAM works, please see the original technical report~\cite{Prisacariu2014} and journal paper~\cite{Kaehler2015}.

\paragraph{Propagation Section}

The propagation section of the pipeline spreads the current label outwards from the existing set of voxels it labels until either there is a significant change in surface normal, or a significant change in colour.
An example of this process is shown in the \emph{Labelling} part of Figure~\ref{fig:workflow}.
Propagation is useful because it allows the user to provide a significant amount of training data to the random forest whilst labelling only a very small part of the scene.
However, it is a comparatively brute force process and must be applied cautiously in order to avoid flooding across weak normal or colour boundaries in the scene.
For more details about how propagation is implemented, see \S\ref{subsec:labelpropagation}.

\paragraph{Training Section}

The training section of the pipeline trains a random forest online over time, using training examples sampled from the current raycast result for each frame.
The first step is to sample similar numbers of voxels labelled with each semantic class (e.g.~chair, floor, table, etc.) from the raycast result (see \S\ref{subsubsec:perlabelvoxelsampling}).
A feature descriptor is then calculated for each voxel (see \S\ref{subsec:featurecalculation}), and paired with the voxel's label to form a training example that can be fed to the random forest.
Finally, the random forest is trained by splitting nodes within it that exhibit high entropy (see \S\ref{subsubsec:foresttraining}).
For more details about the random forest model and the training process, see \S\ref{sec:trainingandprediction}.

\paragraph{Prediction Section}

The prediction section of the pipeline uses the existing random forest to predict labels for currently-unlabelled parts of the scene.
The first step is to uniformly sample voxels from the current raycast result (see \S\ref{subsubsec:uniformvoxelsampling}).
As during training, a feature descriptor is then calculated for each voxel (see \S\ref{subsec:featurecalculation}).
The feature descriptors are then fed into the random forest, which predicts labels for the corresponding voxels.
Finally, these labels are used to mark the voxels in question.
For more details about the prediction process, see \S\ref{subsec:prediction}.

\subsubsection{Rendering}

\stufig{height=12cm}{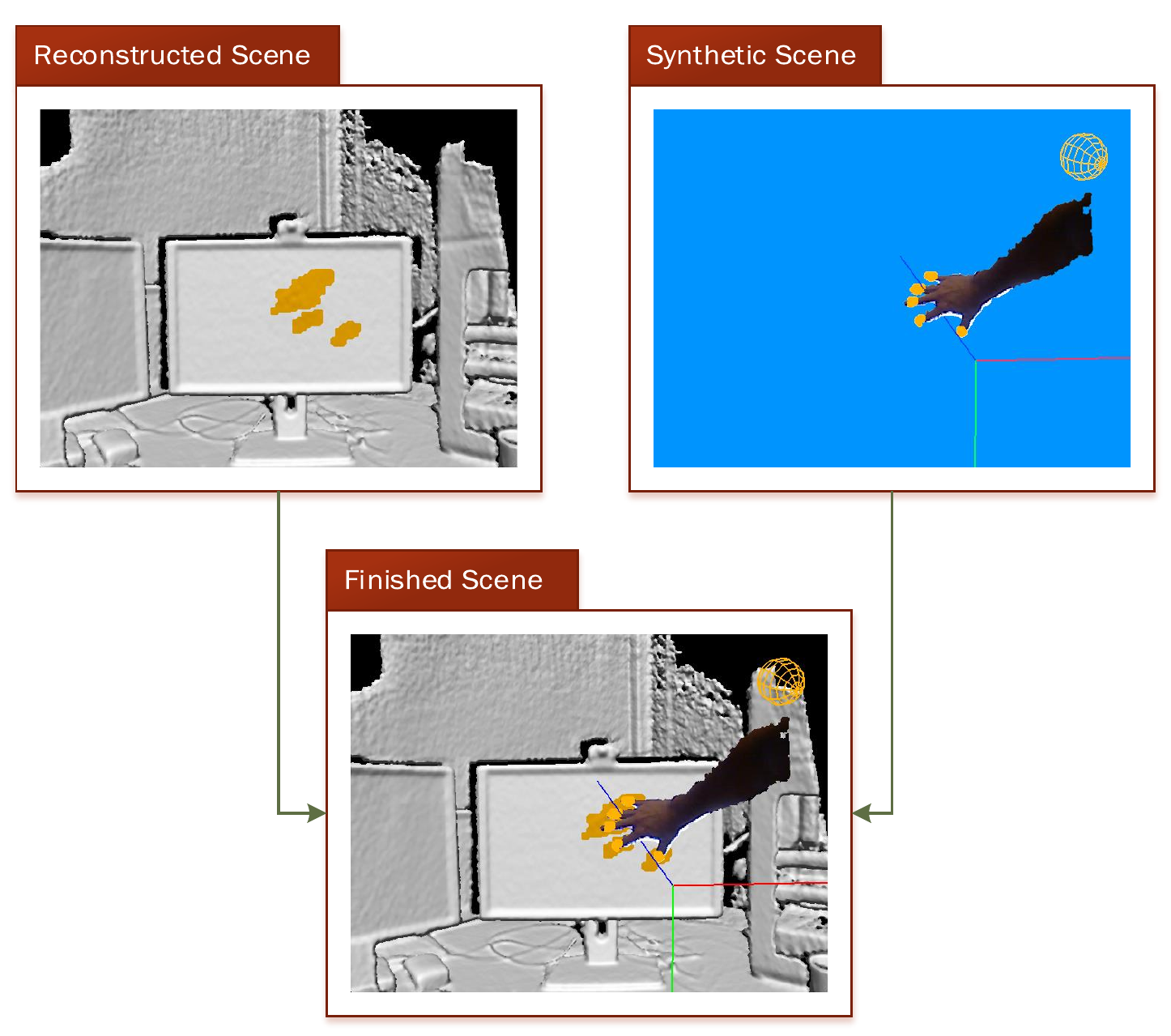}{The compositing of the finished scene.}{fig:rendering}{!t}

The final step in each iteration of the application's main loop is to render the scene.
The \texttt{spaintgui} application currently supports two different renderers: one that performs monocular rendering into a window (which can be made full-screen) and another that performs stereo rendering onto an Oculus Rift head-mounted display.
To support both of these use cases without duplicating any code, we have made it possible to render the scene multiple times from different camera poses.
The windowed renderer then simply renders the scene directly into a window from a single camera pose.
The Rift renderer renders the scene twice (once for each eye) into off-screen textures (using OpenGL's frame buffer capabilities) and passes these textures to the Oculus Rift SDK.

The actual rendering of the scene (see Figure~\ref{fig:rendering}) is split into two parts, the first of which renders a quad textured with a raycast of the reconstructed 3D scene, and the second of which renders a synthetic scene (consisting of the scene axes and the current user interaction tool) over the top of it using OpenGL.
In order to achieve the desired result, the correct model-view and projection matrices must be set when rendering the synthetic scene.\footnotemark{}
The model-view matrix is easily obtained from the current pose of the camera in InfiniTAM.
The projection matrix must be calculated from the intrinsic parameters of the camera being used.

\footnotetext{An additional consideration when rendering on top of a raycasted scene is that the OpenGL depth buffer will not contain correct depth values for the scene in question. If we wanted to render synthetic objects that were correctly occluded by parts of the raycasted scene, we would therefore need to perform a depth raycast of the reconstructed scene and copy it into the OpenGL depth buffer to obtain accurate depth. However, in our case, the objects we want to render do not need to be occluded by the scene, so we are not currently performing this additional step.}

\subsection{The \texttt{rafl} Library}
\label{subsec:rafllibrary}

\stufig{height=23cm}{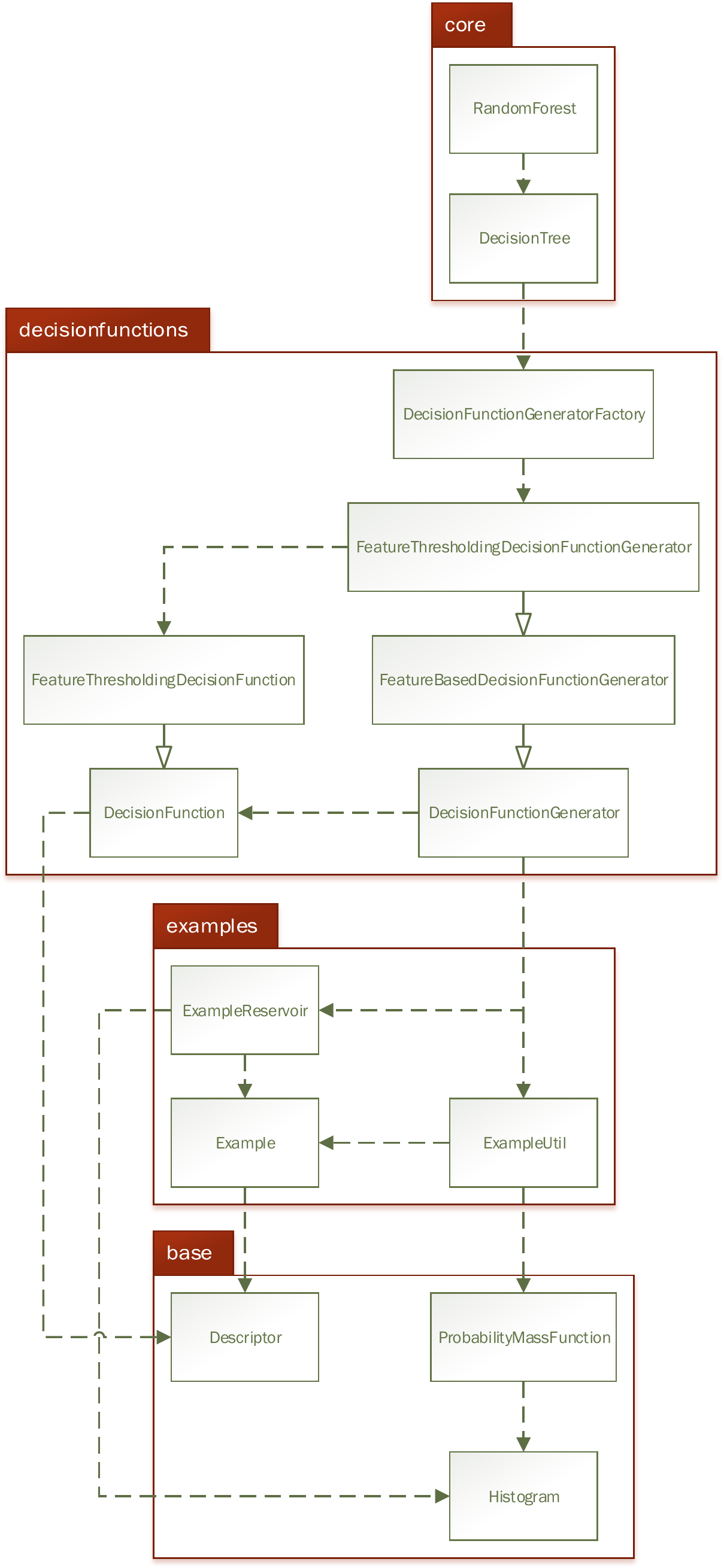}{A simplified version of the \texttt{rafl} architecture. Dashed lines indicate dependencies (in the direction of the arrows); solid lines with open arrowheads indicate inheritance relationships.}{fig:raflarchitecture}{!p}

As previously mentioned, a key part of our \emph{SemanticPaint} framework is the way in which we train a random forest online in order to predict object classes for previously-unlabelled parts of the scene.
Our random forest implementation is provided as a separate library called \texttt{rafl} to maximise its reusability.
For more details about the random forest model, see \S\ref{sec:trainingandprediction}.

As can be seen in Figure~\ref{fig:raflarchitecture}, \texttt{rafl}'s architecture consists of a number of layers.
The lowest layer, \texttt{base}, contains classes that have few dependencies, such as feature descriptors, histograms and probability mass functions.
The next layer, \texttt{examples}, contains classes relating to training examples for random forests.
A training example consists of a feature descriptor and a semantic label.
Each leaf node in the forest stores a `reservoir' of examples -- this is used to keep track of the distribution of examples that have reached that node, whilst limiting the number of examples for each semantic class that are actually stored so as to bound the forest's memory usage.
The \texttt{decisionfunctions} layer contains classes relating to decision functions for random forests.
Each branch node in the forest contains a decision function that can be used to route a feature descriptor down either the left or right subtree of the node.
The decision functions used can be of various types -- for example, a \emph{feature-thresholding} decision function routes feature descriptors by comparing a particular feature in the descriptor to a fixed threshold.
Client code can control the type of decision functions used by a random forest by initialising it with a \emph{decision function generator} that generates decision functions of the desired type.
When a leaf node is split, this decision function generator is used to randomly generate candidate decision functions to split the node's examples.
Finally, the \texttt{core} layer contains the actual decision tree and random forest classes.

\subsection{The \texttt{spaint} Library}

\stufig{height=8cm}{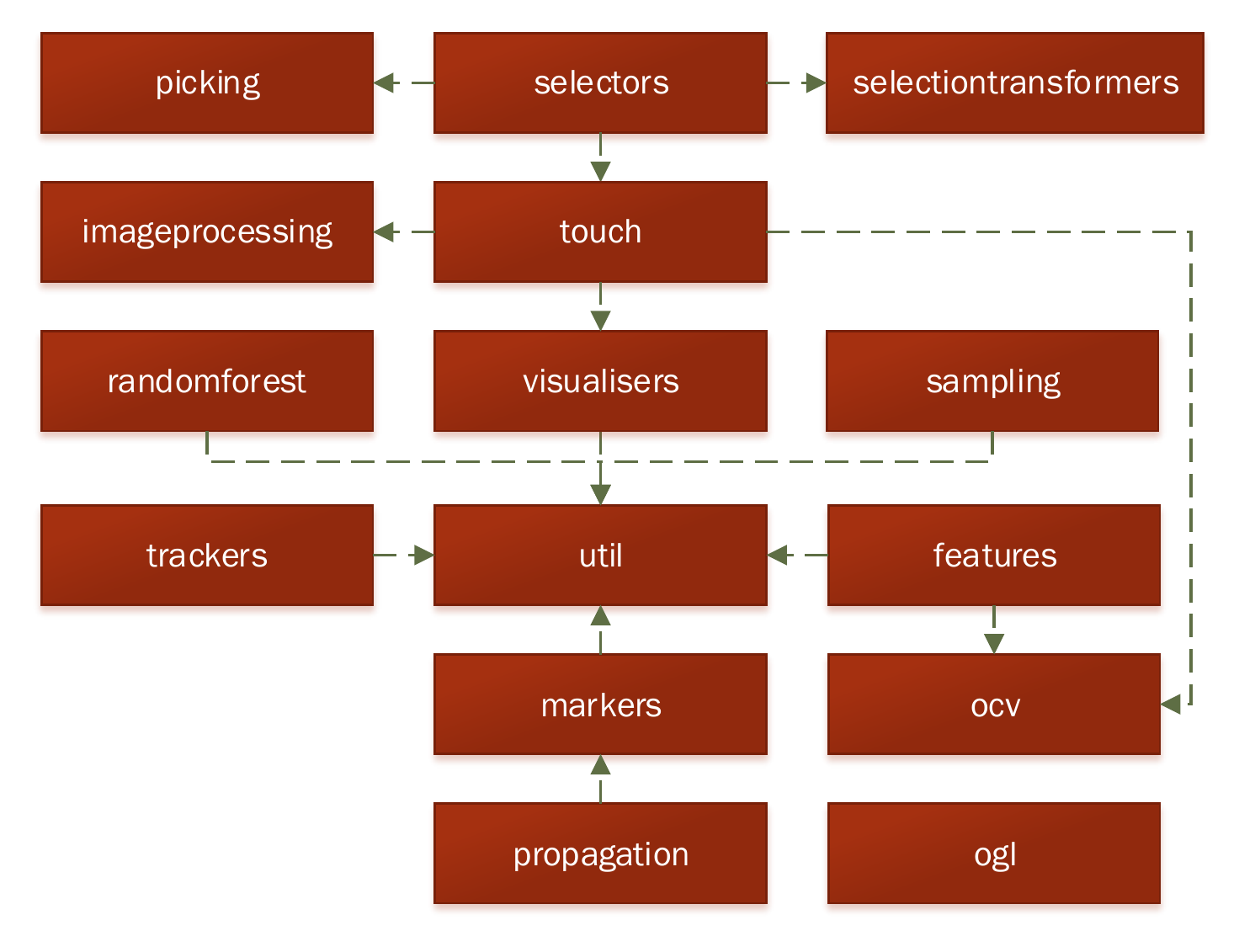}{A component-level view of the \texttt{spaint} architecture. Dashed lines indicate dependencies (in the direction of the arrows).}{fig:spaintarchitecture}{!tp}

The core implementation of \emph{SemanticPaint} is provided in the \texttt{spaint} library.
As can be seen in Figure~\ref{fig:spaintarchitecture}, this is divided into a number of components, the names of which are fairly self-explanatory.
For example, the \texttt{touch} component contains the code to detect when the user physically touches part of the 3D scene.
The bits of the code that sample voxels, calculate features and generate appropriate decision functions for the random forest can be found in \texttt{sampling}, \texttt{features} and \texttt{randomforest} respectively.
The key bits of the code that facilitate user interaction with the scene can be found in \texttt{selectors} and \texttt{selectiontransformers}.
Code to mark voxels with labels is implemented in \texttt{markers}, and code to propagate labels across surfaces can be found in \texttt{propagation}.
Details about how the various components work can be found in the rest of this report.

\section{Labelling the Scene}
\label{sec:labelling}

The primary purpose of interaction in \emph{SemanticPaint} is to allow the user to assign ground truth semantic labels to one or more voxels in the scene.
Users can either label the whole scene manually, or the labels they provide can form a basis for learning models of the various semantic classes (e.g.~table, floor, etc.) that can be used to propagate the ground truth labels to the rest of the scene.

Our framework allows the user to interact with the scene using a variety of modalities, ranging from clicking on individual voxels in a view of the scene using the mouse (known as picking), to physically touching objects in the real world.
In terms of their implementations, these modalities work in very different ways, but at an abstract level they can all be seen as selection generators, or \emph{selectors}: things that generate selections of voxels in the scene.
As a result, we can handle them in a unified way by introducing an abstract interface for selectors, and implementing an individual selector for each modality.

In practice, some modalities (notably picking) initially produce selections that contain too few voxels to provide a useful way of interacting with the scene (it would be tedious for the user to have to label every voxel in the scene individually).
To deal with this problem, we introduce the additional notion of a \emph{selection transformer}, which takes one selection and transforms it into another.
For instance, a simple transformer might expand every voxel in an initial selection into a cube or sphere of voxels to make a more useful selection for labelling.
A more interesting transformer might take account of the scene geometry and spread a selection of voxels along the surface of the scene.
By cleanly separating selection transformation from selection generation, we achieve significant code reuse -- in particular, our selectors can limit themselves to indicating the locations in the scene that the user is trying to select, and delegate the concern of producing a useful selection from that to a subsequent generic transformer.

Once a selection has been generated, it can be used to label the scene.
(It could also be used for other things, e.g.~scene editing, but those are not the focus of the basic framework.)
This can be implemented straightforwardly: the user chooses a label (e.g.~using the keyboard or a voice command) and that label is written into each voxel in the selection.
Figure~\ref{fig:interactionpipeline} shows the overall pipeline we use for interactive scene labelling.

\stufig{height=11.5cm}{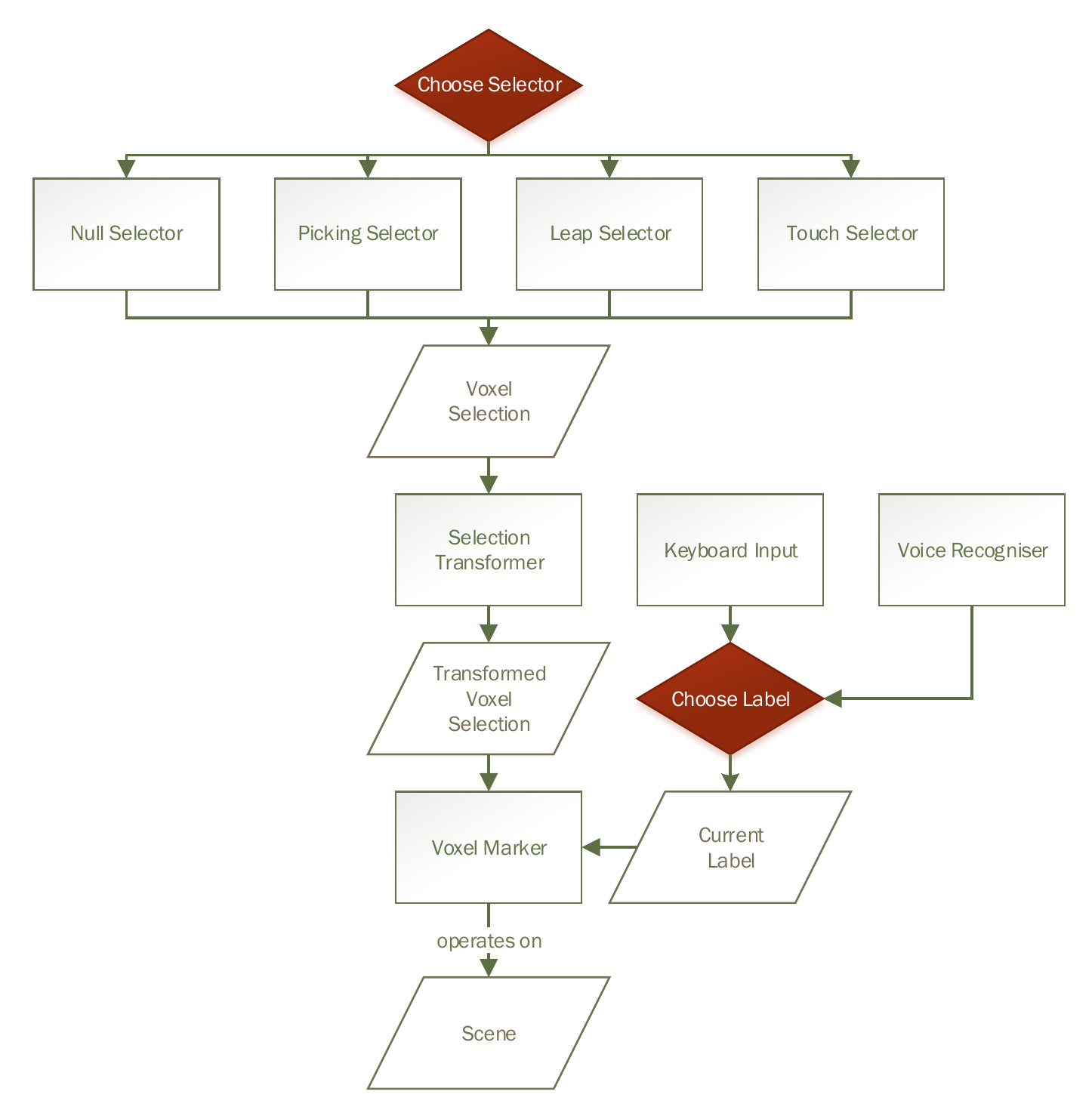}{The pipeline we use for interactive scene labelling.}{fig:interactionpipeline}{!t}

\subsection{Selectors}
\label{subsec:selectors}

The \emph{SemanticPaint} framework incorporates a variety of selectors that the user can use to select voxels in the scene. All of these selectors share a common interface, making it straightforward to add new interaction modalities to the framework. In the following subsections, we describe the two most important selectors that are currently implemented, namely the picking selector and the touch selector.

\subsubsection{Picking Selector}
\label{subsec:pickingselector}

\stufig{width=0.6\textwidth}{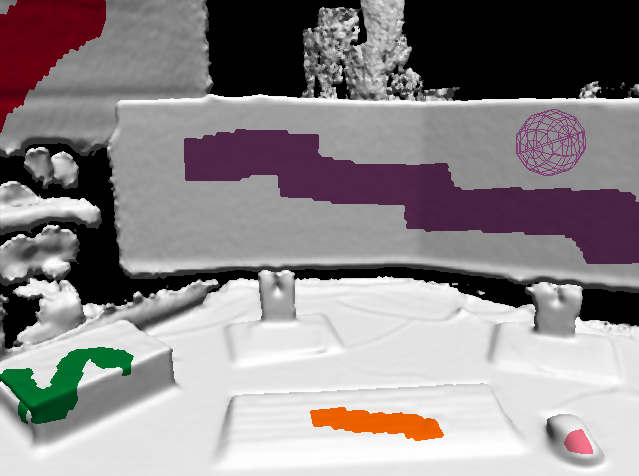}{
An illustration of scene labelling using the picking selector. The coloured orb is used to denote the part of the scene that will be labelled when the user presses the mouse.
}{fig:pickingselector}{!t}

\stufig{width=0.7\textwidth}{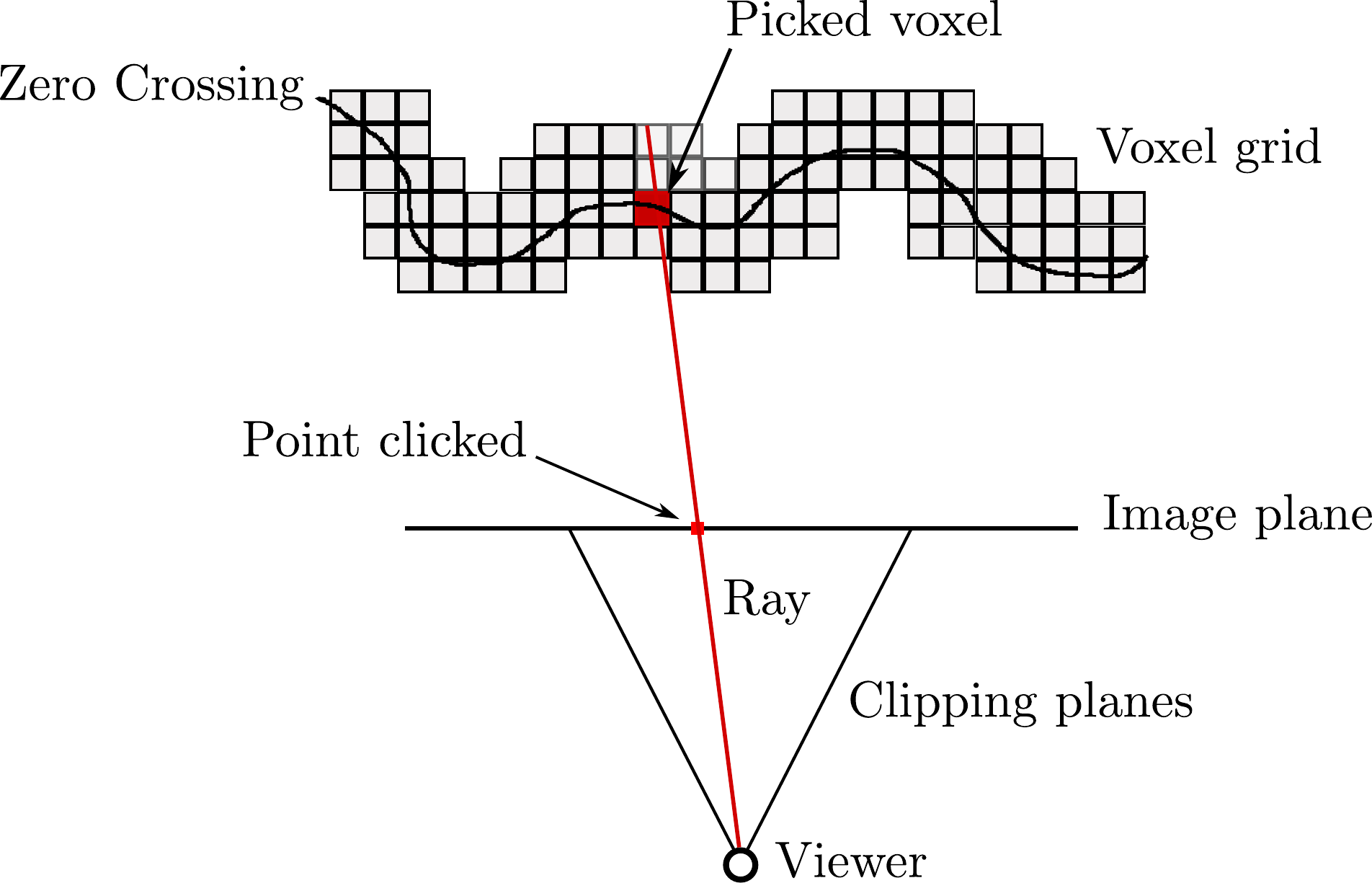}{The picking process.}{fig:picking}{!t}

The picking selector (see Figure~\ref{fig:pickingselector}) allows the user to select a voxel in the scene by clicking a 2D point on the screen with the mouse. This is equivalent to clicking a 3D point on the image plane of the current camera. To determine which voxel the user wants to select, we conceptually cast a ray from the viewer through the clicked point on the image plane and into the scene.\footnotemark{} We then select the first surface voxel (if any) hit by the ray. See Figure~\ref{fig:picking} for an illustration.

The picking selector is simple, but it has a number of advantages: in particular, it not only allows the scene to be labelled precisely, it also allows the user to label distant parts of the scene. Its disadvantage is that it is less intuitive than simply touching scene objects.

\footnotetext{In practice, InfiniTAM already produces a voxel raycast of the scene from the current camera pose at each frame -- this is an image in which each pixel contains the location of the voxel in the scene that would be hit by a ray passing through the corresponding point on the image plane. As a result, the picking selector does not actually need to perform any raycasting itself -- instead, it can simply look up the voxel to select in the voxel raycast.}

\subsubsection{Touch Selector}

The touch selector allows the user to select voxels in the scene by physically touching them with a hand, foot or any other interacting object.
The locations (if any) at which the user is touching the scene are detected by looking at the differences between the live depth image from the camera and a depth raycast of the scene.
An illustration of the touch selector in action can be seen in Figure~\ref{fig:workflow}, and a detailed description of the touch detection process can be found in Appendix~\ref{sec:touchdetection}.

\subsection{Selection Transformers}

As previously mentioned, a \emph{selection transformer} takes one selection of voxels and transforms it into another.
A variety of selection transformers are possible, but at present we have only implemented a simple \emph{voxel-to-cube} transformer, as described in the following subsection.
The first step in the process, for any transformer, is to calculate the size of the output selection as a function of the input selection.
Different transformers can then determine the voxels in the output selection in whichever way they see fit.

\subsubsection{Voxel-to-Cube Transformer}

The \emph{voxel-to-cube} transformer transforms each voxel in the input selection into a fixed-size cube of voxels that is centred on it.
To do this, it first specifies the size of the output selection to be the result of multiplying the size of the input selection by the cube size, and then computes each output voxel on a separate thread (see Listing~\ref{code:voxeltocubetransformer}).

\begin{stulisting}[!t]
\caption{Voxel-to-Cube Transformer}
\label{code:voxeltocubetransformer}
\begin{lstlisting}[style=Default]
/**
 * \brief Writes a voxel to the output selection.
 *
 * \param outputVoxelIndex  The index in the output selection at which to write the voxel.
 * \param cubeSideLength    The length of each side of one of the cubes (in voxels).
 * \param cubeSize          The number of voxels in each cube.
 * \param radius            The (Manhattan) radius (in voxels) to select around each
 *                          initial voxel.
 * \param inputSelection    The input selection of voxels.
 * \param outputSelection   The output selection of voxels.
 */
_CPU_AND_GPU_CODE_
inline void write_voxel_to_output_selection(int outputVoxelIndex, int cubeSideLength,
  int cubeSize, int radius, const Vector3s *inputSelection, Vector3s *outputSelection)
{
  // Look up the input voxel that corresponds to this output voxel (there is one input
  // voxel for each cube of output voxels).
  int inputVoxelIndex = outputVoxelIndex / cubeSize;
  const Vector3s& inputVoxel = inputSelection[inputVoxelIndex];

  // Calculate the linear offset of the output voxel within its cube.
  int offset = outputVoxelIndex % cubeSize;

  // Use this linear offset to calculate the 3D offset of the output voxel from the
  // position of the input voxel.
  int xOffset = offset % cubeSideLength - radius;
  int yOffset = offset % (cubeSideLength * cubeSideLength) / cubeSideLength - radius;
  int zOffset = offset / (cubeSideLength * cubeSideLength) - radius;

  // Write the output voxel to the output selection.
  outputSelection[outputVoxelIndex] = Vector3s(
    inputVoxel.x + xOffset,
    inputVoxel.y + yOffset,
    inputVoxel.z + zOffset
  );
}
\end{lstlisting}
\end{stulisting}

\subsection{Voxel Marking}

When a user selects some voxels and chooses a label with which to mark them, actually marking the voxels with the label in question is straightforward. However, manual labelling by the user is not the only way in which voxels can acquire a label: voxels can also be labelled either during label propagation or as the result of predictions by the random forest. Since it is important to ensure that `reliable' labels provided by the user are not overwritten by such automatic processes, this creates the need for a way of distinguishing between labels based on their provenance -- in particular, \emph{SemanticPaint} currently stores a \emph{group} identifier alongside each label, allowing it to distinguish between user labels, propagated labels, and labels predicted by the forest.\footnote{In practice, we bit-pack a label and its group into a single $8$-bit byte in each voxel to save space.}

The voxel marker uses these group identifiers to control the labelling process and prevent undesirable overwriting of user labels. More specifically, it considers the old label of a voxel and its proposed new label, and marks the voxel with the new label if and only if either (a) both the old and new labels are non-user labels, or (b) the new label is a user label. One minor complication is that this scheme does not interact well with the undo system -- without special handling, it is not possible to undo the overwriting of a non-user label with a user label, since the voxel marker prevents the overwriting of user labels with non-user labels. For this reason, in practice the voxel marker has a special `force' mode that is used when undoing a previous labelling -- this bypasses the normal checks and allows any label to overwrite another.

\subsection{Label Propagation}
\label{subsec:labelpropagation}

The overall purpose of our system is to make it possible for users to label a 3D scene with a minimum of effort, and to this end, we train a random forest classifier, based on the labelled examples the user provides, that can be used to predict labels for the whole scene.
However, users tend to label the scene quite sparsely. Whilst we can certainly train a forest using this sparse data, and even achieve reasonable results, we would naturally expect that much better results can be obtained by training from more of the scene.
In practice, therefore, we augment the training examples provided by the user using a process called \emph{label propagation} before training the random forest.

Label propagation operates on the current raycast result for each frame.
It works by deciding, for each voxel in the raycast result, whether or not to set its label to the current semantic label, based on the labels, positions, colours (in the CIELab space) and normals of neighbouring voxels in the raycast result image.\footnotemark{}
A voxel's label will be updated if some of its neighbouring voxels have the current semantic label and are sufficiently close in position, colour and normal to the voxel under consideration.
Over a number of frames, this has the effect of gradually spreading the current semantic label across the scene until position, colour or normal discontinuities are encountered.
The implementation of this process is shown in Listing~\ref{code:labelpropagation}, and an illustration showing a label gradually propagating across a scene can be found in Figure~\ref{fig:workflow}.

Although label propagation is effective in augmenting the training examples provided by the user, it does have some limitations.
The most notable of these is that because it is essentially performing a flood fill on the scene, it is vulnerable to weak property boundaries.
Whilst we mitigate this to some extent by using a conjunction of different properties that are less likely to all have a weak boundary in the same place in the scene, it is still entirely possible for propagation to spread the current semantic label into parts of the scene that the user did not intend.
We control this in two ways: first, by spreading the label relatively slowly across the scene, we allow the user to stop the propagation at an appropriate point; second, we allow the user to revert back to the sparse labels they originally provided and try the propagation again.
In practice, we find that this gives users a reasonable degree of control over the propagation and allows them to make effective use of it to speed up their labelling of the scene.

\footnotetext{In practice, we do not use the \emph{direct} neighbours of each voxel in the image, but we elide over this implementation detail to simplify the explanation.}

\begin{stulisting}[!p]
\caption{The label propagation process}
\label{code:labelpropagation}
\begin{lstlisting}[style=Default]
/* Determines whether or not the specified label should be propagated from a
   specified neighbouring voxel to the target voxel. */
inline bool should_propagate_from_neighbour(int neighbourX, int neighbourY,
  int width, int height, SpaintVoxel::Label label,
  const Vector3f& loc, const Vector3f& normal, const Vector3u& colour,
  const Vector4f *raycastResult, const Vector3f *surfaceNormals,
  const SpaintVoxel *voxelData, const ITMVoxelIndex::IndexData *indexData,
  float maxAngleBetweenNormals, float maxSquaredDistBetweenColours,
  float maxSquaredDistBetweenVoxels)
{
  if(neighbourX < 0 || neighbourX >= width || neighbourY < 0 || neighbourY >= height)
	  return false;

  // Look up the position, normal and colour of the neighbouring voxel.
  int neighbourVoxelIndex = neighbourY * width + neighbourX;
  Vector3f neighbourLoc = raycastResult[neighbourVoxelIndex].toVector3();

  bool foundPoint;
  SpaintVoxel neighbourVoxel =
    readVoxel(voxelData, indexData, neighbourLoc.toIntRound(), foundPoint);
  if(!foundPoint) return false;

  Vector3f neighbourNormal = surfaceNormals[neighbourVoxelIndex];
  Vector3u neighbourColour = VoxelColourReader::read(neighbourVoxel);

  // Compute the angle between the neighbour's normal and the target voxel's normal.
  float angleBetweenNormals =
    acosf(dot(normal, neighbourNormal) / (length(normal) * length(neighbourNormal)));

  // Compute the distance between the neighbour's colour and the target voxel's colour.
  Vector3f colourOffset = rgb_to_lab(neighbourColour) - rgb_to_lab(colour);
  float squaredDistBetweenColours = dot(colourOffset, colourOffset);

  // Compute the distance between the neighbour's position and the target voxel's
  // position.
  Vector3f posOffset = neighbourLoc - loc;
  float squaredDistBetweenVoxels = dot(posOffset, posOffset);
	float distBetweenVoxels = sqrt(squaredDistBetweenVoxels);

  // Decide whether or not propagation should occur.
  return neighbourVoxel.packedLabel.label == label &&
         angleBetweenNormals <= maxAngleBetweenNormals * distBetweenVoxels &&
         squaredDistBetweenColours <= maxSquaredDistBetweenColours * distBetweenVoxels &&
         squaredDistBetweenVoxels <= maxSquaredDistBetweenVoxels;
}

/* Propagates the specified label to the specified voxel as necessary, based on its own
   properties and those of its neighbours. */
inline void propagate_from_neighbours(int voxelIndex, int width, int height,
  SpaintVoxel::Label label, const Vector4f *raycastResult, const Vector3f *surfaceNormals,
  SpaintVoxel *voxelData, const ITMVoxelIndex::IndexData *indexData,
  float maxAngleBetweenNormals, float maxSquaredDistBetweenColours,
  float maxSquaredDistBetweenVoxels)
{
  // Look up the position, normal and colour of the specified voxel.
  Vector3f loc = raycastResult[voxelIndex].toVector3();
  Vector3f normal = surfaceNormals[voxelIndex];

  bool foundPoint;
  const SpaintVoxel voxel = readVoxel(voxelData, indexData, loc.toIntRound(), foundPoint);
  if(!foundPoint) return;

  Vector3u colour = VoxelColourReader::read(voxel);

  // Based on these properties and the properties of the neighbouring voxels, decide if
  // the specified voxel should be marked with the propagation label, and mark it if so.
  int x = voxelIndex % width;
  int y = voxelIndex / width;

#define SPFN(nx,ny) should_propagate_from_neighbour(nx, ny, /* common parameters */)

  if((SPFN(x - 2, y) && SPFN(x - 5, y)) ||
     (SPFN(x + 2, y) && SPFN(x + 5, y)) ||
     (SPFN(x, y - 2) && SPFN(x, y - 5)) ||
     (SPFN(x, y + 2) && SPFN(x, y + 5)))
  {
    mark_voxel(loc.toShortRound(), PackedLabel(label, LG_PROPAGATED), /* ... */);
  }

#undef SPFN
}
\end{lstlisting}
\end{stulisting}

\section{Learning the Scene}
\label{sec:trainingandprediction}

In this section we briefly describe how we can train a random forest model to predict the probability that a specified voxel takes a particular object label.
We use random forests because they have been employed successfully in many computer vision applications \cite{Breiman2001,Criminisi2012} and are extremely fast at making predictions.
Moreover, they can be trained online \cite{Valentin2015} and support the addition of new object categories on the fly.

\subsection{Random Forests}
\label{subsec:randomdecisiontrees}
A binary decision tree is a classifier that can be used to map vectors in a feature space $\Set{X}$ to one of a number of discrete labels in a finite label space $\Set{Y}$.\footnotemark{}
Each branch node in the tree contains a decision function $\MathFunction{d}: \Set{X} \mapsto \{L,R\}$ that can be used to send feature vectors down either the left ($L$) or right ($R$) subtree of the node.
Each leaf node contains a \emph{probability mass function} $\MathFunction{p}: \Set{Y} \mapsto [0,1]$ that satifies $\sum_{y \in \Set{Y}} \MathFunction{p}(y) = 1$ and specifies a probability $\MathFunction{p}(y)$ for each label $y \in \Set{Y}$.

\footnotetext{Note that in general, a decision tree may support arbitrary label spaces $\Set{Y}$ such as structured output label spaces \cite{Dollar2013,Kontschieder2011}, since at test time the leaf node reached only depends on the input vector $\Vector{x}$.}

Starting from the root node, we can use such a tree to classify a feature vector by repeatedly testing it against the decision function in the current branch node and then recursing down either the left or right subtree as appropriate.
The classification process terminates when a leaf node is reached, at which point an $\argmax$ over the probability mass function in the leaf can be used as the classification result.

Binary decision trees are generally trained by selecting decision functions at the branch nodes that are good at dividing the available training examples into different classes.
Since the number of different decision functions that could be chosen for a branch node is generally infinite, it is common to randomly generate a moderate, finite number of candidate decision functions for a node and pick one that best divides the available data.
This heuristic approach is not guaranteed to find a globally optimal decision function, but will generally give a reasonable split of the examples in practice.

One way of improving the classification performance of a system based on binary decision trees is to combine multiple such trees in an ensemble called a \emph{random forest} and perform classification by averaging the probability mass functions they yield for each given feature vector (the final label for a vector can be computed as an $\argmax$ over this average). The individual trees in the forest are trained on the same examples, but will have different structures because of the way in which the candidate decision functions are randomly generated during training. Training a random forest rather than an individual tree reduces the propensity of the system to overfit to the training data and allows it to better generalise to new examples.

One factor that affects the computational efficiency and classification performance of a decision tree is the specific form of decision function used at the branch nodes. For \emph{SemanticPaint}, we adopted simple `stump' decision functions of the form described in \cite{Criminisi2012}, which work by testing a single feature $x_k$ (i.e.~the $\Index{k}\ith$ feature in a feature vector $\Vector{x}$) against a threshold $\tau$:
\begin{equation}
\label{eqn:decisionfunction}
  \MathFunction{d}(\Vector{x}) = \phi_{\Index{k},\tau}(\Vector{x}) = \begin{cases}
L & \mbox{if } x_k < \tau \\
R & \mbox{otherwise}
\end{cases}
\end{equation}
Decision functions of this form are attractive because they are extremely efficient to evaluate whilst still supporting good overall classification performance.

\subsection{Online Learning}
\label{subsec:streamingdecisionforests}
In order to allow the random forest to learn from an incoming stream of training examples 
and to update itself as new data arrives,
we use the `streaming decision forest' algorithm described in \cite{Valentin2015}.\footnote{Note that `decision forest' is a synonym for `random forest'.}
The main idea is to use reservoir sampling \cite{Vitter1985} to maintain an unbiased sample of examples of fixed maximum size at each leaf node.
The use of reservoirs bounds the memory usage (allowing larger trees to be grown) in a way that still ensures that there is an equal chance of keeping examples in the reservoir over time.
Further details of the algorithm may be found in \cite{Valentin2015}.

\subsection{Feature Calculation}
\label{subsec:featurecalculation}

The random forests we use in \emph{SemanticPaint} learn to semantically segment a scene based on features computed from local geometry and colour information around individual voxels. In this section, we describe the features we use (originally developed in \cite{Valentin2015}), and the way in which they are calculated. The basic requirements we have for features are as follows:
\begin{itemize}
\item They should capture the local geometric and colour context around voxels.
\item They should be very fast to compute (since we have to compute them for thousands of voxels per frame).
\item They should be viewpoint-invariant.
\item They should be at least somewhat illumination-invariant.
\end{itemize}
To meet these requirements, \cite{Valentin2015} introduced \emph{voxel-oriented patch} (VOP) features.
These primarily consist of an $n \times n$ colour patch (where $n$ defaults to $13$) around the voxel, extracted from the tangent plane to the surface at that point and rotated to align with the dominant orientation in the patch.
The colours are expressed in the CIELab colour space to achieve some level of illumination invariance.
The full feature descriptor also includes the surface normal at the voxel to make it easier to classify key elements in the scene such as the floor and walls.

\stufig{width=\linewidth}{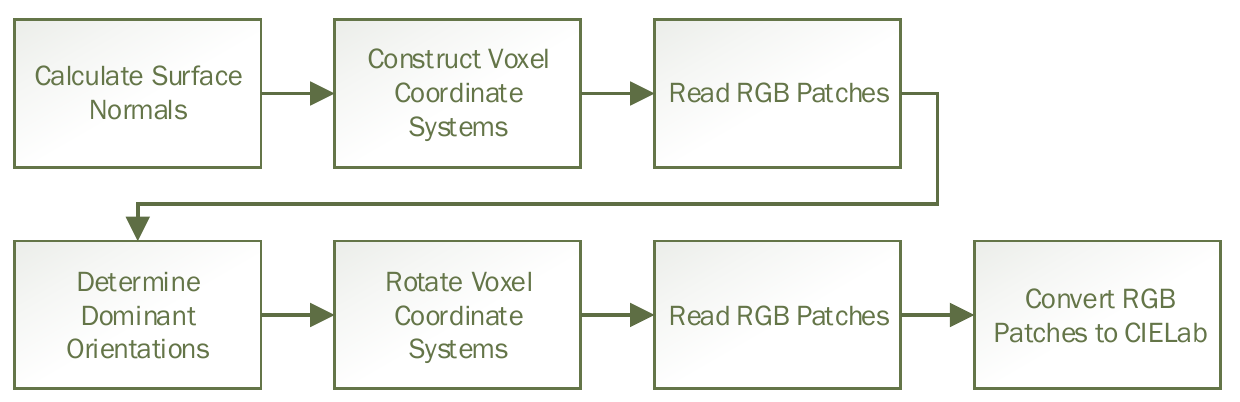}{The feature calculation pipeline.}{fig:featurecalculation}{!t}

The pipeline we use for computing features is shown in Figure~\ref{fig:featurecalculation}.
The first step is to calculate surface normals for all the target voxels for which we want to compute features.
We use these to construct an arbitrary 2D coordinate system in the tangent plane to the surface at each voxel.\footnotemark{}
For each target voxel, we then imagine placing a grid aligned with the voxel's coordinate system over the scene, and read an $n \times n$ RGB patch from the voxels surrounding the target voxel on that grid.
Next, we compute a histogram of gradients for each of these patches, and use it to find the dominant gradient orientation around each voxel.
We then rotate the coordinate system for each voxel to align it with the voxel's dominant gradient orientation, and read another $n \times n$ RGB patch from the voxel's new coordinate system.
Finally, the RGB patches in this second batch are converted to the CIELab colour space and combined with the surface normals to form the full feature descriptors.

\footnotetext{The method we use to generate an arbitrary 2D coordinate system in a plane is described in \S{}G.3 of \cite{Golodetz2006}.}

\subsection{Training}
\label{subsec:training}

Having looked at how we can compute features for individual voxels, we are now in a position to describe how we train our random forests.
The training process is online and spread across multiple frames. At each frame, we sample a number of voxels from the reconstructed scene.
For each such voxel, we then construct a training example and add it to the forest.
Finally, the random forest is trained by choosing leaf nodes in the various trees to split based on their entropy.
In the following subsections, we discuss in more detail the way in which we sample voxels from the scene and the way in which we use training examples constructed from the sampled voxels to train the forest.

\subsubsection{Sampling Voxels}
\label{subsubsec:perlabelvoxelsampling}

\stufig{height=24cm}{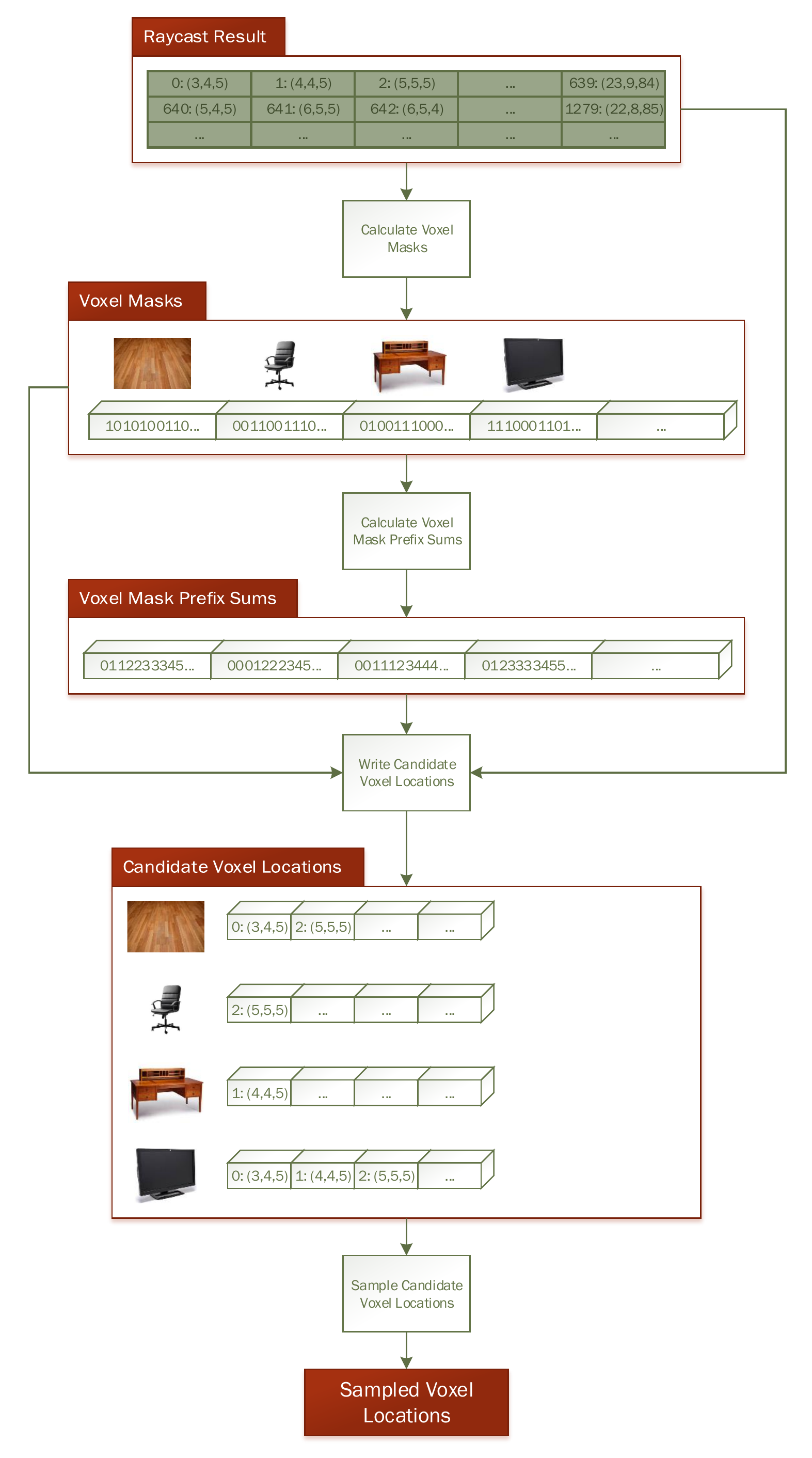}{The workings of the per-label voxel sampler, which is used to pick voxels with which to train the random forest (see \S\ref{subsubsec:perlabelvoxelsampling}).}{fig:perlabelvoxelsampler}{!p}

To acquire training data with which to train the random forest, we randomly choose voxels from the scene to serve as examples of particular semantic labels.
These voxels are passed to the feature calculator (see \S\ref{subsec:featurecalculation}), which in turn produces feature descriptors that can be used to form training examples.
The input to the training sampler is the current voxel raycast of the scene: that is, at each frame we sample from the voxels that are visible from the live camera position.

The goal of the training sampler is to produce relatively even numbers of voxels from the different semantic classes so as to train a forest that does not unfairly prioritise one class over another.
To do this, it first groups the voxels in the current voxel raycast by semantic label to form a set of candidate voxels for each label, and then randomly chooses similar numbers of voxels from the candidate sets to produce a final set of sampled voxels for each label.

Although this is conceptually quite a simple process, in practice the implementation is complicated by the need to sample voxels as quickly as possible to achieve real-time frame rates.
To do this, we parallelise the entire process, as illustrated in Figure~\ref{fig:perlabelvoxelsampler}.
The first step is to produce a \emph{voxel mask} for each semantic class that is in use: the mask for a class contains a bit for each voxel in the current raycast result, denoting whether or not that voxel is labelled with the class in question.
Having generated the masks, the goal is to write the voxels for each class whose mask bits are set to $1$ into an array of candidate voxels for that class from which voxels can then be randomly sampled for training purposes.
To allow these voxels to be written into the candidate voxel arrays in parallel, however, the threads doing the writing need to know the array locations into which to write the voxels.
To calculate these locations, we resort to a standard parallel programming trick.
Starting from the voxel masks, we first efficiently compute their prefix sums, as described in \cite{Harris2007}: these tell us the locations into which to write the voxels.
Each thread then writes its voxel into the candidate voxel array at the specified location if and only if the corresponding mask bit is set to $1$.
Having generated arrays containing candidate voxels for each semantic class, the only thing that remains is to then sample from them randomly to produce the final output.

\subsubsection{Training the Forest}
\label{subsubsec:foresttraining}
Having sampled voxels for each semantic class, it is then straightforward to make training examples from them with which to train the forest.
To do this, we simply calculate a feature descriptor for each sampled voxel (see \S\ref{subsec:featurecalculation}) and pair each feature descriptor with its voxel's label to make a training example.

These training examples can then be added to the forest, a process that involves separately adding all of the examples to each decision tree.
To add an example to a tree (see Listing~\ref{code:addexample}), we simply pass it down the tree to an appropriate leaf node by testing its feature descriptor against the decision functions in the branch nodes it encounters on the way.
Once it reaches a leaf, we add it to the leaf's example reservoir (as mentioned in \S\ref{subsec:streamingdecisionforests}, the use of reservoirs bounds the memory usage of the forest by limiting the number of examples that are explicitly stored in each leaf).

\begin{stulisting}[!t]
\caption{Adding an example to a decision tree}
\label{code:addexample}
\begin{lstlisting}[style=Default]
/**
 * \brief Adds a new training example to the decision tree.
 *
 * \param example The example to be added.
 */
void add_example(const Example_CPtr& example)
{
  // Find the leaf to which to add the new example.
  int leafIndex = find_leaf(*example->get_descriptor());

  // Add the example to the leaf's reservoir.
  m_nodes[leafIndex]->m_reservoir.add_example(example);

  // Mark the leaf as dirty to ensure that its splittability is properly recalculated
  // once all of the examples have been added.
  m_dirtyNodes.insert(leafIndex);

  // Update the class frequency histogram.
  m_classFrequencies.add(example->get_label());
}

/**
 * \brief Finds the index of the leaf to which an example with the specified descriptor
 *        would currently be added.
 *
 * \param descriptor  The descriptor.
 * \return            The index of the leaf to which an example with the descriptor
 *                    would currently be added.
 */
int find_leaf(const Descriptor& descriptor) const
{
  int curIndex = m_rootIndex;
  while(!is_leaf(curIndex))
  {
    if(m_nodes[curIndex]->m_splitter->classify_descriptor(descriptor) == DC_LEFT)
      curIndex = m_nodes[curIndex]->m_leftChildIndex;
    else
      curIndex = m_nodes[curIndex]->m_rightChildIndex;
  }
  return curIndex;
}
\end{lstlisting}
\end{stulisting}

Training the forest involves splitting its leaf nodes so as to separate the examples they contain based on their semantic classes.
Given a set of examples $\Set{S}$ that have accumulated in a particular leaf node, the task of splitting the node is to find a decision function that splits those examples into sets $\Set{S}_L$ and $\Set{S}_R$, typically in a way that maximises the amount of information gained in the process.
(Intuitively, the idea is that a `good' decision function will try and separate examples with distinct class labels.)
The leaf can then be replaced with a branch node containing the decision function, and two child nodes that each contain some of the examples from the old leaf node.

In practice, it is typical to choose a decision function from among a number of different candidates, scored using a cost function that encourages the average entropy of the example distributions in the new child nodes to be lower than the entropy of the distribution in the original leaf.
Formally, the score of the split induced by a particular decision function can be calculated using the information gain objective
\begin{equation}
  \label{eqn:informationgain}
  \MathFunction{g}\left(\Set{S},\Set{S}_{L},\Set{S}_{R}\right) = \MathFunction{h}(\Set{S})
  - \sum_{b \in \{L,R\}} \frac{\left|\Set{S}_{b}\right|}{\left|\Set{S}\right|} \MathFunction{h}(\Set{S}_{b}),
\end{equation}
in which $\MathFunction{h}(\Set{S})$ denotes the Shannon entropy of the distribution of class labels in $\Set{S}$.

Candidate decision functions are generated randomly, generally from a parameterised family of functions with a predefined structure.
In our case, we are using decision functions that test an individual feature $x_k$ against a threshold $\tau$ (see \S\ref{subsec:randomdecisiontrees}), so we can generate candidate decision functions simply by randomly generating pairs $(\Index{k}_i, \Scalar{\tau}_i)$ that determine the functions in question.
When splitting a leaf, we first generate a moderate number of candidate decision functions, and then choose a candidate that maximises the information gain in (\ref{eqn:informationgain}) as the decision function for the new branch node.

We have seen how to split an individual leaf node, but the question of how to schedule such splits so as to train the overall forest remains.
In practice, since we are trying to train our forest online whilst maintaining a real-time frame rate, a convenient way to implement the scheduling process is to maintain a priority queue of nodes to be split for each tree in the forest.
We assign each node in each tree a \emph{splittability} score and use the priority queues to rank them in non-increasing order of splittability.
The splittability score of a node with example set $\Set{S}$ is defined to be
\begin{equation}
  \Scalar{s} =
  \begin{cases}
    \MathFunction{h}(\Set{S}) & \mbox{if } \left| \Set{S} \right| \geq \alpha, \\
    0 & \mbox{otherwise,}
  \end{cases}
\end{equation}
in which $\alpha$ is a tunable constant. This captures the intuitions that (a) nodes with a higher number of examples will contain more accurate probability mass functions, and (b) nodes with a higher entropy should be split first, since they are more likely to achieve a high information gain when split.
At each frame, we split nodes in each tree, starting from the most splittable, until we either run out of nodes to split or exceed a specified split budget.
This process allows us to focus attention on the nodes we are most keen to split whilst maintaining a real-time frame rate.

\subsection{Prediction}
\label{subsec:prediction}

Having trained a random forest model for the scene from sparse user labels as described in \S\ref{subsec:training}, we would like to use it to predict semantic labels for the remaining voxels in the scene.
In practice, however, there are far too many voxels in the scene as a whole to simultaneously predict labels for all of them in real time.
Instead, we adopt a strategy of gradually predicting labels for voxels over time.
At each frame, we randomly sample a proportion of the voxels in the current view\footnotemark{} and predict their labels; over time, as the camera moves around the scene, this results in us labelling all of the surface voxels in the scene.
In the following subsections, we first describe how we sample voxels from the scene, and then describe how we use the random forest to predict labels for the sampled voxels.

\footnotetext{The current default is $8192$ voxels, but this is configurable.}

\subsubsection{Sampling Voxels}
\label{subsubsec:uniformvoxelsampling}

By contrast with the process of sampling voxels for training, where it is important to sample balanced numbers of voxels from each semantic class, the sampling process for prediction is extremely straightforward.
Indeed, since we are primarily trying to predict labels for currently-unlabelled voxels in the scene, one might think it would suffice simply to randomly sample unlabelled voxels from the current raycast result.
In practice, we simplify things even further by allowing labelled voxels to be predicted more than once: this reduces the process to one of uniformly sampling voxels with replacement from the current raycast result, which can be straightforwardly implemented by generating random pixel indices on the CPU.
Moreover, this scheme allows previously-labelled voxels to have their predictions updated as the forest changes over time.

The slight drawback of this approach is that it can take some time to fill in the last few unlabelled voxels in the current view.
For the purposes of visual feedback, this can be mitigated by applying median filtering when rendering the scene to remove impulsive noise.
For the purposes of further computation, however, it would be better to focus attention on the unlabelled voxels in the scene before repredicting previously-labelled voxels: we plan to improve the sampler to account for this in future iterations of our framework.

\subsubsection{Predicting Voxel Labels}

To predict a label for each sampled voxel, we first calculate a feature descriptor for it as described in \S\ref{subsec:featurecalculation}.
We pass this descriptor down each decision tree in the forest until we reach a leaf, using the \texttt{find\_leaf} function shown in Listing~\ref{code:addexample}.
Each of the leaves found by this process contains a probability mass function specifying the probabilities that the voxel should be labelled with each of the various semantic classes that are being used to label the scene.
To determine an appropriate label for the voxel, we first average these probability distributions to produce a probability mass function for the forest as a whole, and then deterministically choose a label with the greatest probability.
An illustration of how label prediction works can be seen in Figure~\ref{fig:prediction}, and the corresponding implementation can be found in Listing~\ref{code:predictlabel}.

\stufig{height=8cm}{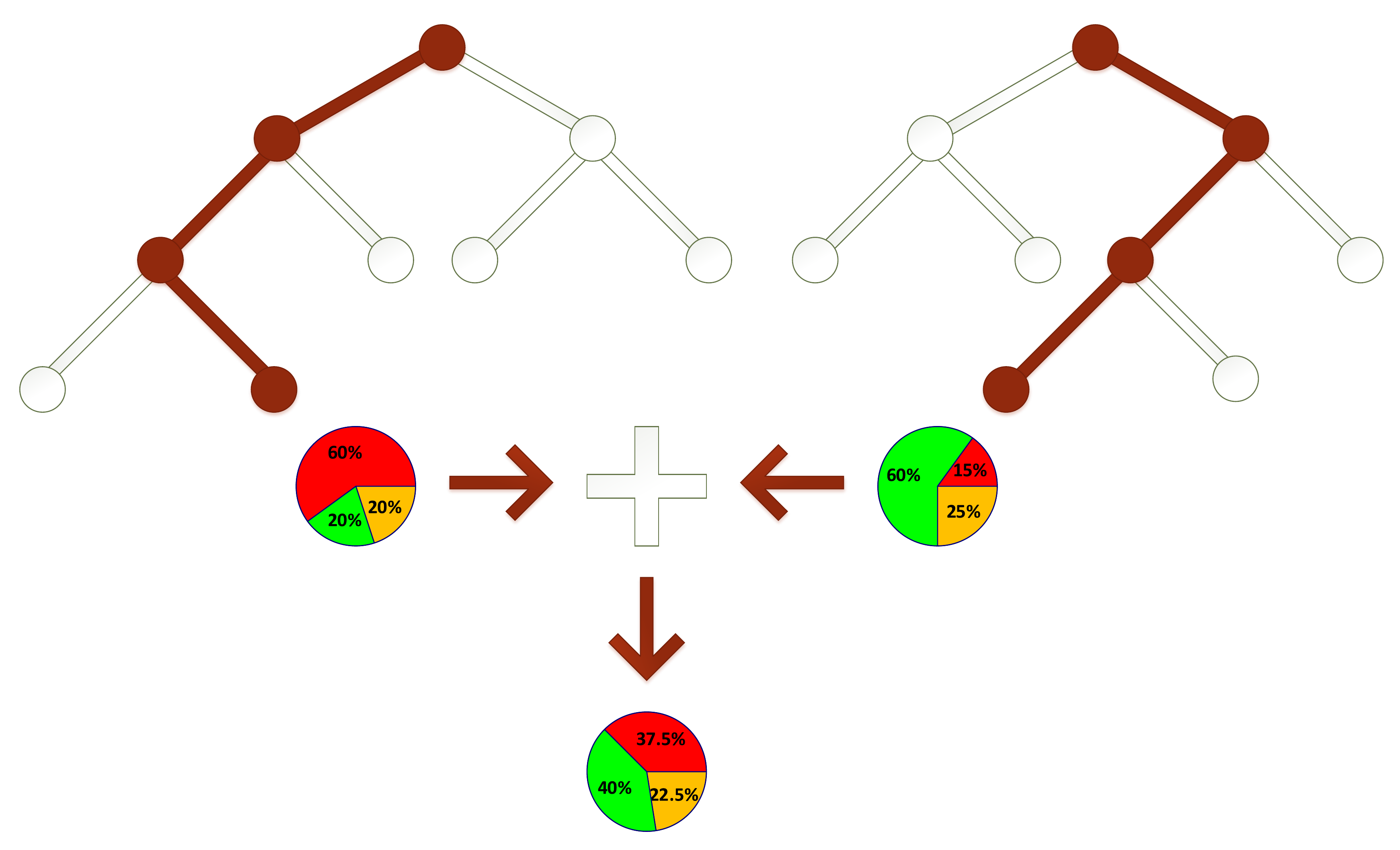}{The label prediction process. The feature vector for which we want to predict a label is passed down each tree in the forest separately, yielding a probabiliy mass function for each tree. These are averaged to produce a probability mass function for the forest as a whole, and the final label is predicted by performing an $\argmax$ over the average distribution.}{fig:prediction}{!p}

\begin{stulisting}[!p]
\caption{Predicting the label to assign to a voxel in the scene}
\label{code:predictlabel}
\begin{lstlisting}[style=Default]
/**
 * \brief Calculates an overall forest PMF for the specified descriptor.
 *
 * This is simply the average of the PMFs for the specified descriptor in the various
 * decision trees.
 *
 * \param descriptor  The descriptor.
 * \return            The PMF.
 */
ProbabilityMassFunction<Label> calculate_pmf(const Descriptor_CPtr& descriptor) const
{
  // Sum the masses from the individual tree PMFs for the descriptor.
  std::map<Label,float> masses;
  for(auto it = m_trees.begin(), iend = m_trees.end(); it != iend; ++it)
  {
    ProbabilityMassFunction<Label> individualPMF = (*it)->lookup_pmf(descriptor);
    const std::map<Label,float>& individualMasses = individualPMF.get_masses();
    for(auto jt = individualMasses.begin(), jend = individualMasses.end();
		    jt != jend; ++jt)
    {
      masses[jt->first] += jt->second;
    }
  }

  // Create a normalised probability mass function from the summed masses.
  return ProbabilityMassFunction<Label>(masses);
}
	
/**
 * \brief Predicts a label for the specified descriptor.
 *
 * \param descriptor  The descriptor.
 * \return            The predicted label.
 */
Label predict(const Descriptor_CPtr& descriptor) const
{
  return calculate_pmf(descriptor).calculate_best_label();
}
\end{lstlisting}
\end{stulisting}

\subsection{Preliminary Quantitative Results}
The multi-class classification performance of our random forest library was evaluated on the UCI Poker dataset \cite{Lichman2013}, since it contains $10$ categories with a large imbalance in the number of samples per class,
as often encountered in the real-world usage of \emph{SemanticPaint}.
For instance, it is common for the user to point the camera at certain parts of the scene for longer than others,
which results in gathering more samples for some classes than others.

With the exact same data splits and evaluation procedure as described in \cite{Shotton2013}, we achieved an average accuracy of $63.86 \pm 0.44 \%$ (better than $50.121\%$, the
accuracy achieved by always predicting the `nothing in hand' class).
However, since the prior distribution of the categories in Poker is highly non-uniform, the percentage of correctly-classified examples may not be an appropriate measure of overall
classification performance \cite{Everingham2010}.
After normalising the confusion matrix to account for the imbalance in the distribution of classes, the multi-class accuracy was $13.87 \pm 0.09\%$ (random chance is $10\%$).

In the presence of unbalanced data, the multi-class prediction accuracy may be improved by weighting each category $\Scalar{z} \in \Set{Y}$ with the inverse class frequencies
observed in the training data $\Set{D}$, $\Scalar{w}_\Scalar{z} = \left(\sum_{(\Vector{x},\Scalar{y}) \in \Set{D}} \left[\Scalar{y} = \Scalar{z} \right] \right)^{-1}$.
Moreover, these weights may be considered  in the computation of the (weighted) information gain \cite{Kontschieder2011}.
We found that reweighting the probability mass functions during learning and prediction \cite{Kontschieder2011} doubled the normalised classification accuracy on Poker to $26.18 \pm 0.87\%$.

\section{Example Scenes}
\begin{figure}[!t]
  \centering
  \includegraphics[width=0.98\textwidth]{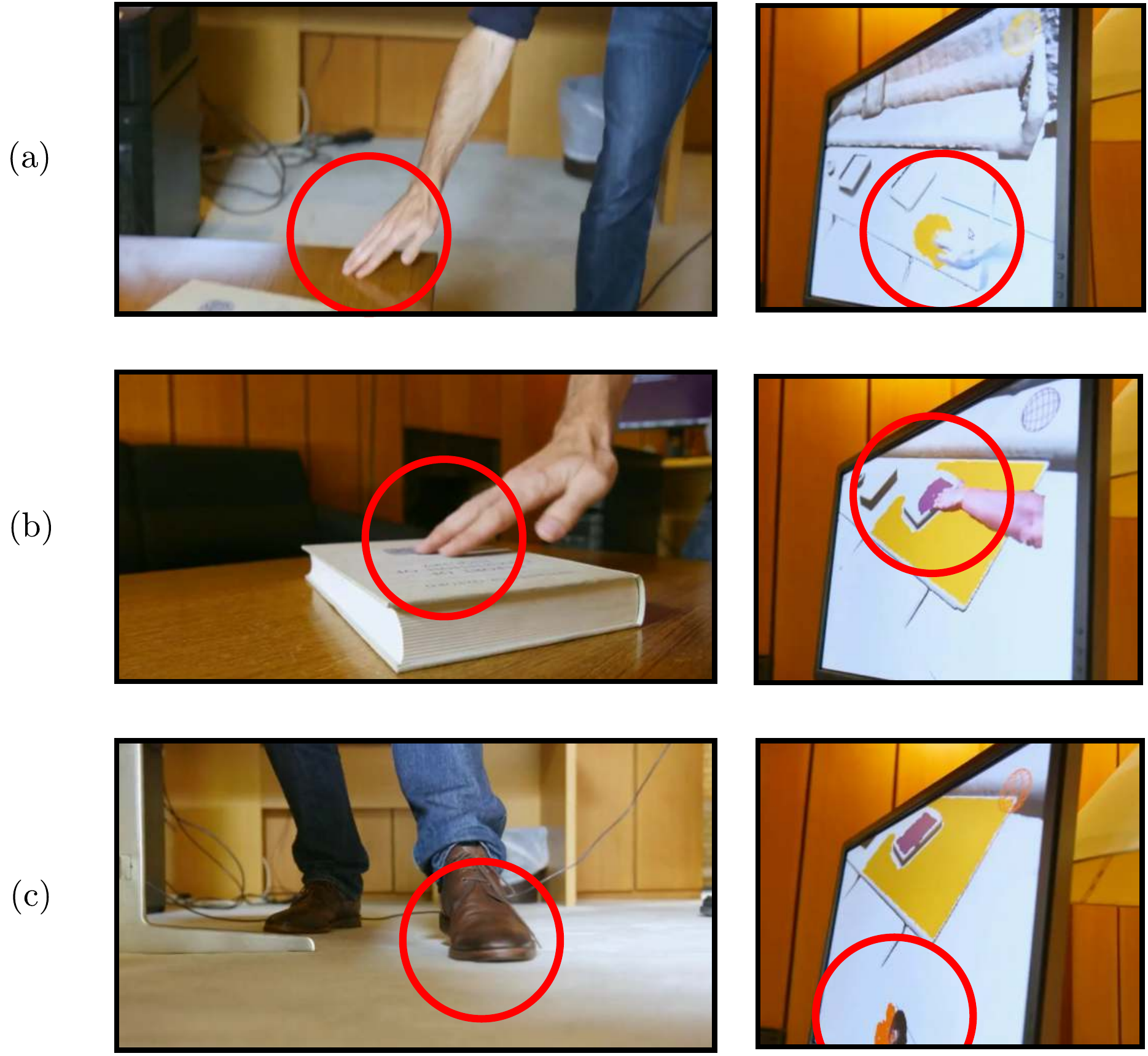}
  \caption{
    Examples of touch interaction and label propagation being used to interactively label a 3D scene.
    \textbf{(a)} A user labels the coffee table by touching it with his hand.
    \textbf{(b)} The labelling in (a) has been propagated across the surface of the table up to colour and normal discontinuities.
    The user now labels the book using a different semantic class.
    \textbf{(c)} Finally, the user labels the floor using his foot.
  }
  \label{fig:examplescenescatstouchprop}
\end{figure}

\begin{figure}[!t]
  \centering
  \includegraphics[width=0.98\textwidth]{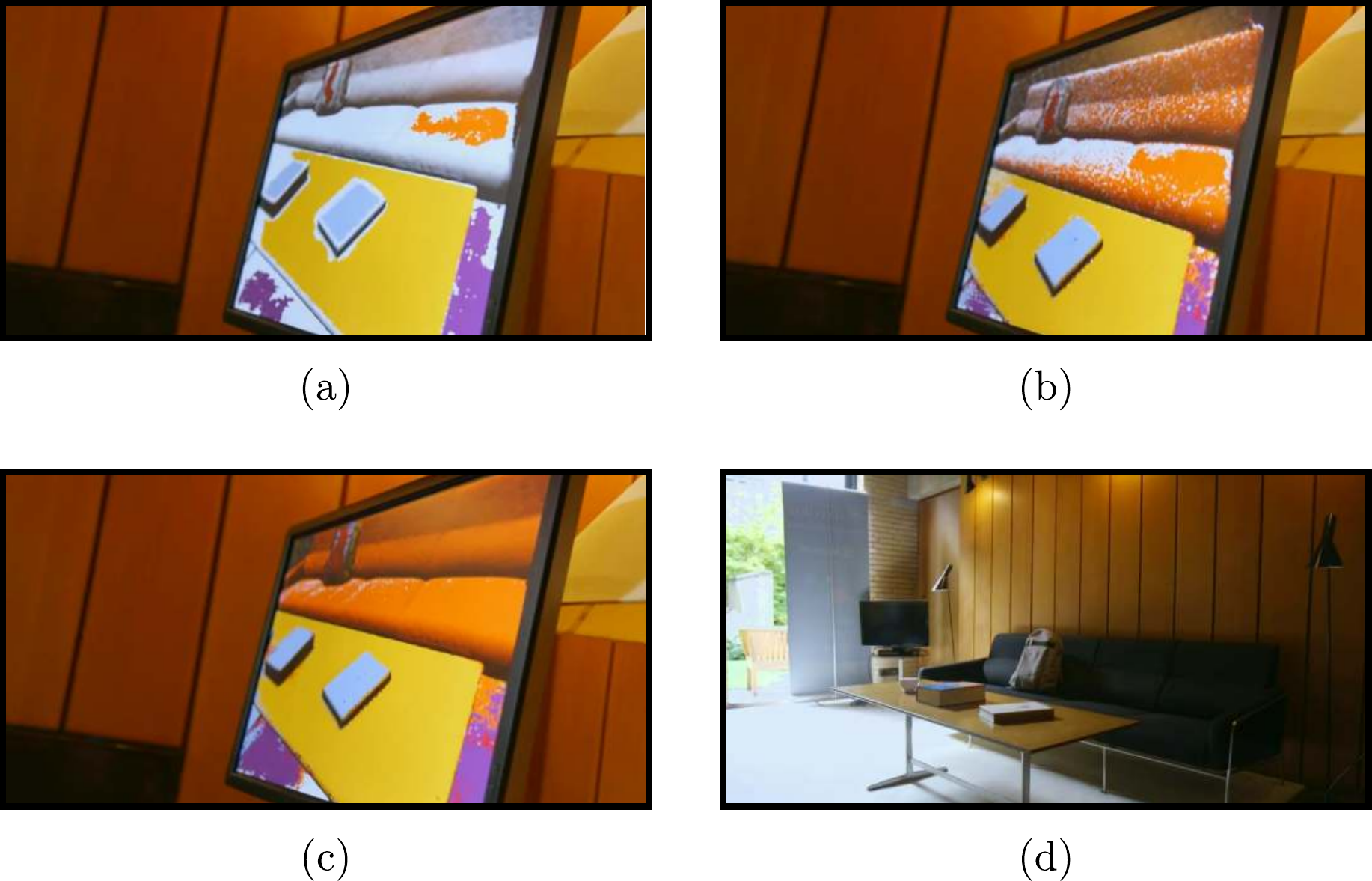}
  \caption{
    An illustration of the prediction step detailed in \S\ref{subsec:prediction} on a re-run of the system on the scene shown in Figure~\ref{fig:examplescenescatstouchprop}.
    \textbf{(a)} The scene after initial labelling by the user and label propagation.
    \textbf{(b)-(c)} The prediction process gradually fills in the labels for the rest of the scene.
    \textbf{(d)} The actual scene being labelled.
  }
  \label{fig:examplesscenescatsresults}
\end{figure}

Having discussed some of the details of how our framework is implemented,
we now present a few qualitative examples to illustrate its performance on some real-world scenes.
In Figure~\ref{fig:examplescenescatstouchprop}, we show the user interacting with the 3D world in order to assign labels to object categories such as `coffee table', `book' and `floor'.
Note that after initially labelling the scene using touch interaction, the labels are propagated across smooth surfaces in the scene as described in \S\ref{subsec:labelpropagation}.
After training the random forest on the propagated user labels, the labels of the remaining voxels in the scene are predicted in real time by the forest,
as illustrated in Figure~\ref{fig:examplesscenescatsresults}.

\section{Conclusion}

In this report, we have described \emph{SemanticPaint} \cite{Valentin2015,Golodetz2015}, our interactive segmentation framework for 3D scenes, built on top of the InfiniTAM v2 3D reconstruction engine of K{\"a}hler \emph{et al.} \cite{Kaehler2015}. Using our framework, a user can quickly segment a reconstructed 3D scene by providing `ground truth' semantic labels at sparse points in the scene and then taking advantage of label propagation and an online random forest model to efficiently label the rest of the scene.

Our framework supports a number of different modalities for user interaction, ranging from selecting scene voxels using the mouse to physically touching objects in the real world. Moreover, it can be extended straightforwardly to support new modalities.

By keeping the user `in the loop' during the segmentation process, we allow any initial classification errors made by the random forest to be quickly corrected. This is important, because automatic segmentation is in general an ill-posed problem: different users want different results, so in the absence of user input it is unclear what the desired output of the segmentation process should look like.

From an architectural perspective, our \emph{SemanticPaint} framework has been implemented as a set of portable libraries that currently work on Windows, Ubuntu and Mac OS X. Although we have not currently done so, it would be reasonably straightforward to port them to further platforms if needed. We very much hope that they can form the basis for further research in this area.

\appendix

\section{Camera Control}
\label{sec:cameracontrol}

\stufig{height=7cm}{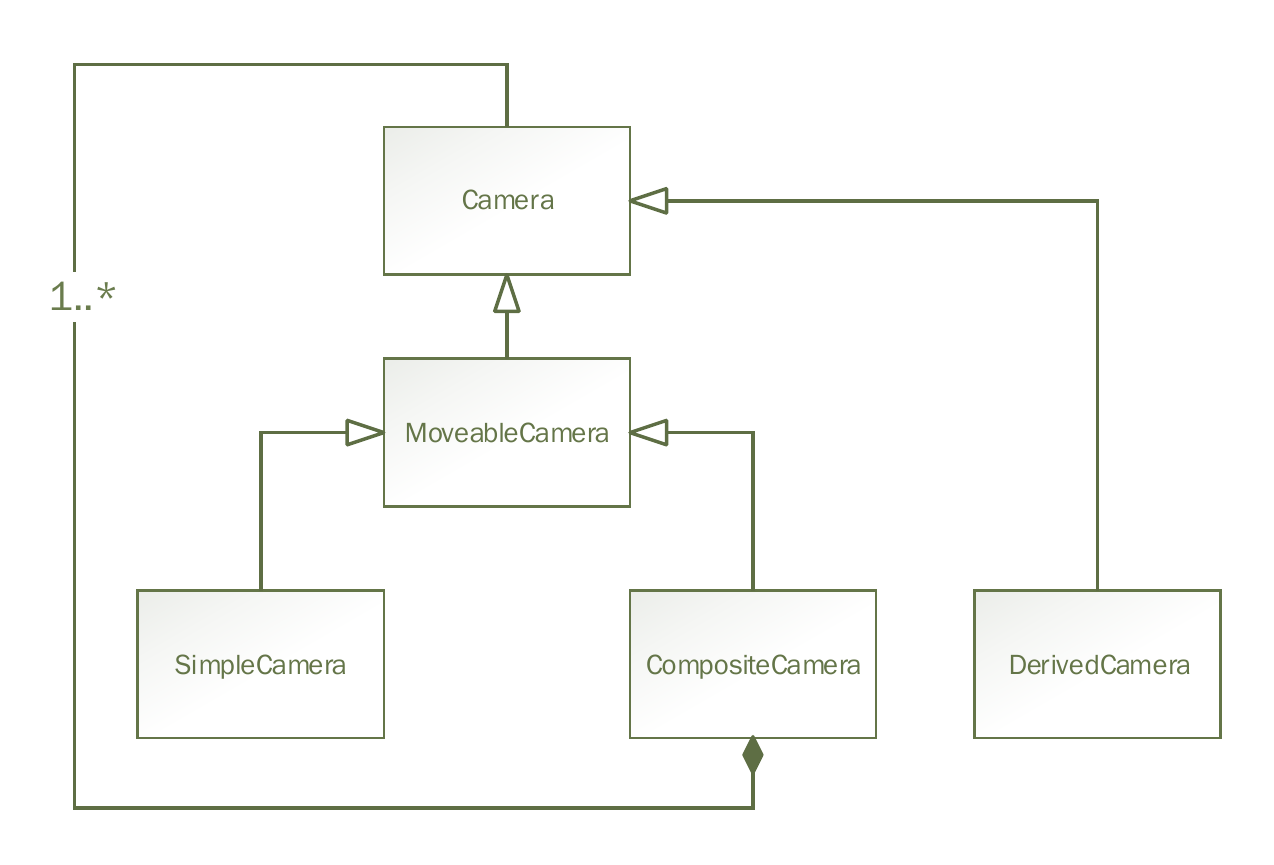}{The architecture of the \texttt{rigging} library.}{fig:riggingarchitecture}{!p}

\stufig{height=5cm}{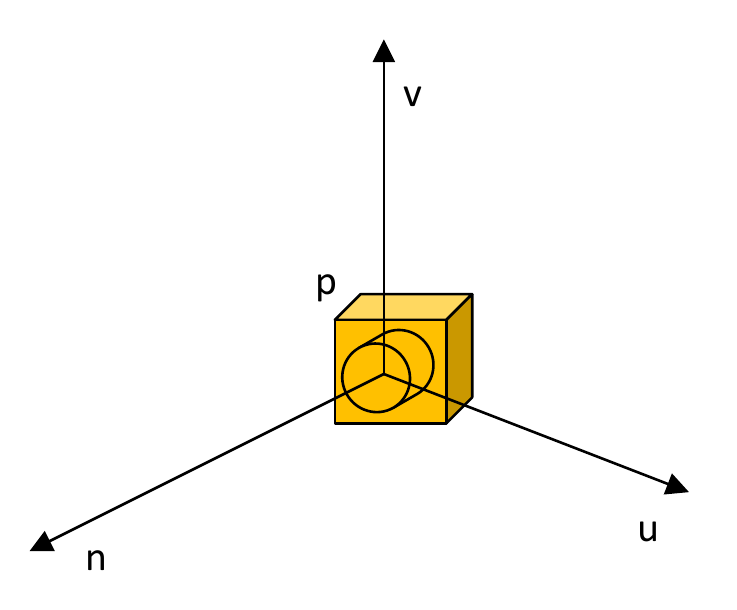}{The \texttt{Camera} class in the \texttt{rigging} library represents a camera in 3D space using its position ($\mathbf{p}$) and a set of orthonormal axes as shown.}{fig:cameraaxes}{!p}

\stufig{height=7cm}{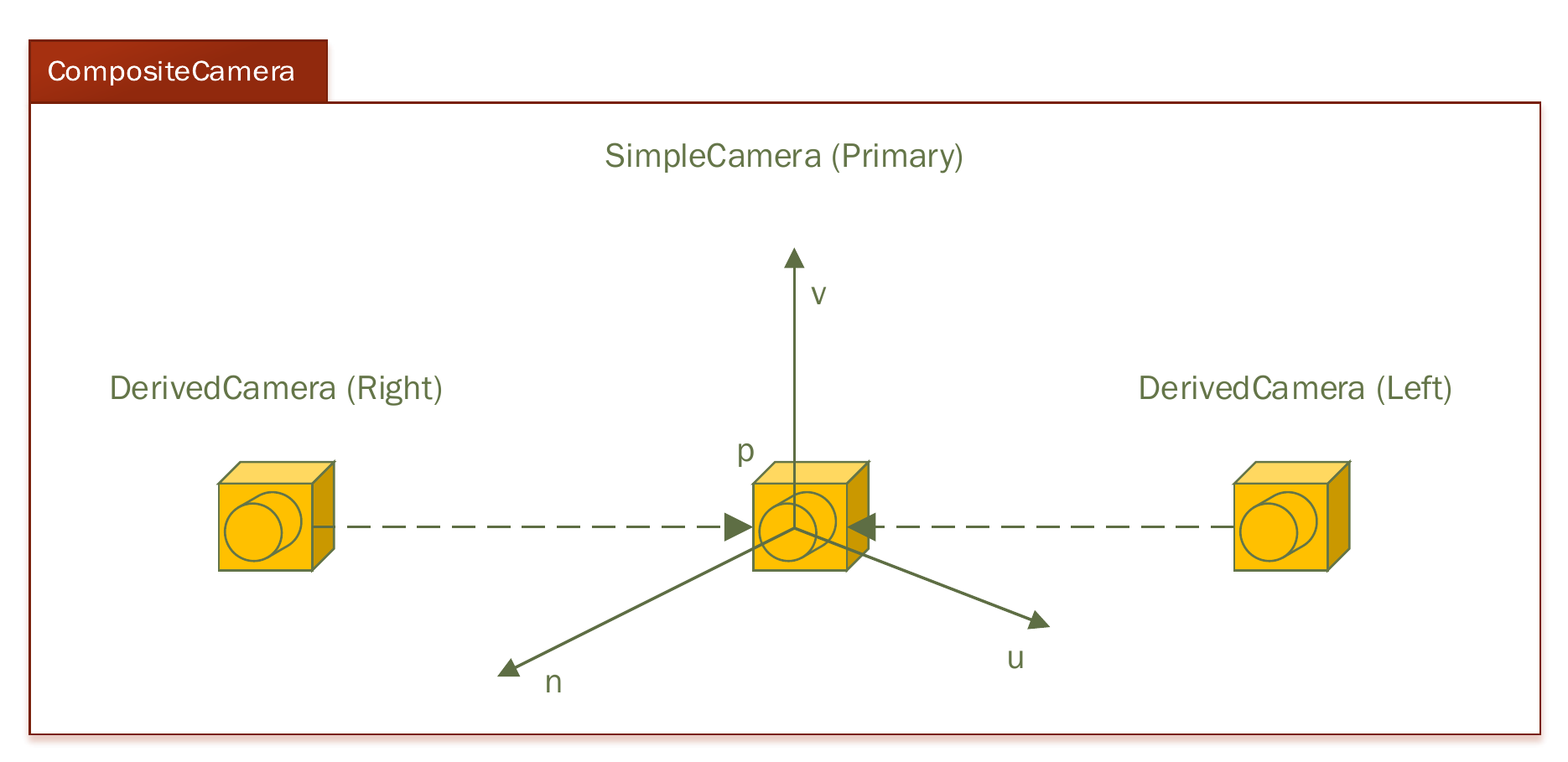}{The way in which the \texttt{rigging} classes can be used together to represent a simple stereo rig. Moving the composite moves the primary camera, which in turn moves the secondary cameras that are derived from it.}{fig:stereorig}{!p}

Camera control in \emph{SemanticPaint} is implemented in the \texttt{rigging} library.
This contains a hierarchy of camera classes, as shown in Figure~\ref{fig:riggingarchitecture}.
The abstract \texttt{Camera} class, at the root of the hierarchy, represents a camera in 3D space using its position and a set of orthonormal axes (see Figure~\ref{fig:cameraaxes}).
The abstract \texttt{MoveableCamera} class derives from this, adding member functions that allow a camera to be translated and rotated in space.
There are then three concrete camera classes: \texttt{SimpleCamera}, \texttt{DerivedCamera} and \texttt{CompositeCamera}.
A \texttt{SimpleCamera} is a concrete implementation of \texttt{MoveableCamera} that stores explicit position and axis vectors.
A \texttt{DerivedCamera} is a non-moveable camera whose position and orientation are based on those of a \emph{base} camera: it stores a rotation and translation in camera space, and its position and orientation are computed on-demand relative to the position and orientation of the base camera.
This is useful, because it allows rigid camera rigs to be represented in a straightforward way.
Finally, a \texttt{CompositeCamera} is a moveable rig of several cameras: it consists of a single \emph{primary} camera that controls the position and orientation of the rig itself, and a number of secondary cameras that are generally directly or indirectly based on that camera.
An example showing how these classes can be used together to represent a simple stereo rig can be seen in Figure~\ref{fig:stereorig}.

\section{Command Management}
\label{sec:commandmanagement}

\stufig{height=8cm}{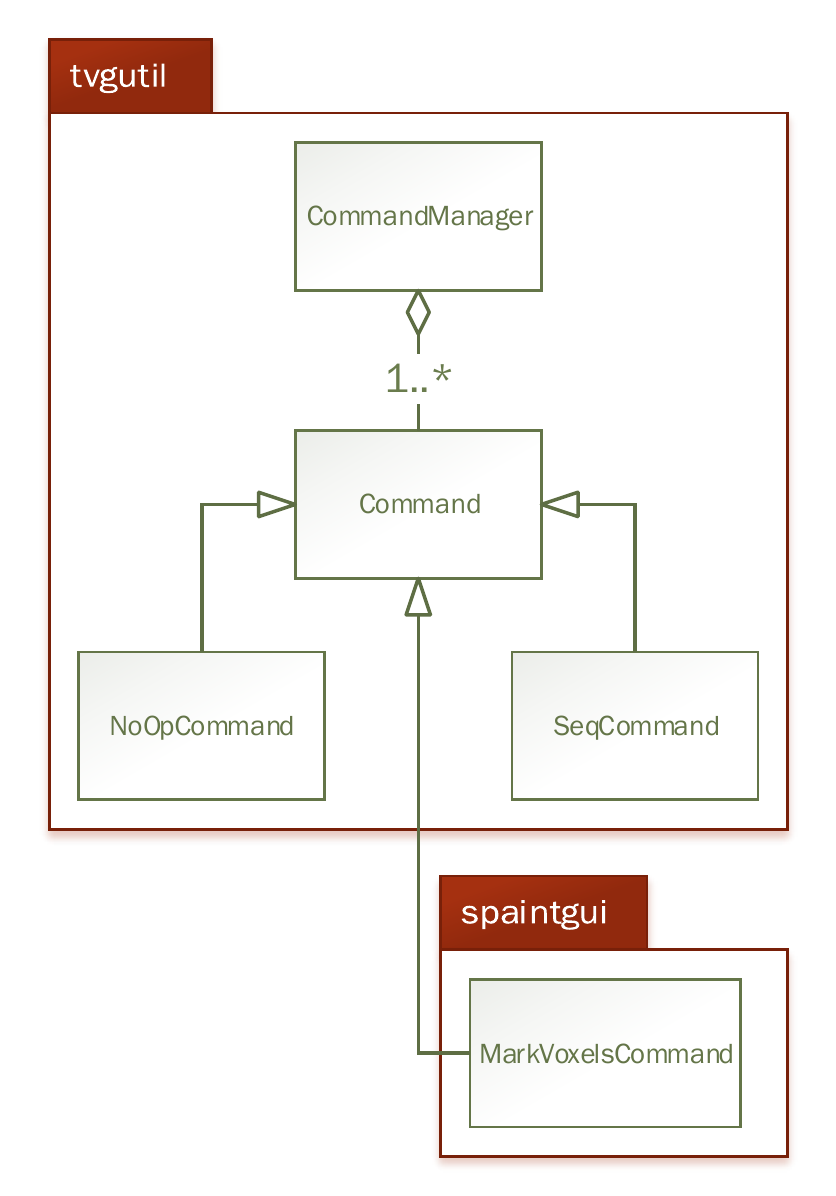}{The architecture of the command system.}{fig:commandsystem}{!t}

To make it possible for users to undo and redo their actions, \emph{SemanticPaint} incorporates a command system of the form described in \cite{Golodetz2006} and originally based on \cite{Sufrin2004}.
We briefly discuss the workings of this system here for completeness.\footnote{We borrow from \cite{Golodetz2006} for this purpose, with the permission of the original author.}

The command system is primarily implemented in the \texttt{tvgutil} library.
As illustrated in Figure~\ref{fig:commandsystem}, it consists of classes relating to the commands themselves, and a command manager.

\subsection{Commands}

The root of the command class hierarchy is an abstract base class called \texttt{Command}.
The interface of this class allows commands to be both executed and undone (see Listing~\ref{code:command}).

\begin{stulisting}[!t]
\caption{The \texttt{Command} class}
\label{code:command}
\begin{lstlisting}[style=Default]
/**
 * \brief An instance of a class deriving from this one can be used to represent a command
 *        that can be executed/undone/redone.
 */
class Command
{
  //...

protected:
  explicit Command(const std::string& description);

public:
  virtual ~Command();

public:
  /** \brief Executes the command. */
  virtual void execute() const = 0;

  /** \brief Undoes the command. */
  virtual void undo() const = 0;

public:
  /** \brief Gets a short description of what the command does. */
  const std::string& get_description() const;
};
\end{lstlisting}
\end{stulisting}

Atomic actions in an application using this system can be represented using classes that derive directly from \texttt{Command} and implement \texttt{execute} and \texttt{undo} member functions.
Composite actions (those that involve executing a sequence of smaller actions) can be represented using an instance of \texttt{SeqCommand}, a helper class whose \texttt{execute} implementation executes a known sequence of commands in order, and whose \texttt{undo} implementation undoes the same commands in reverse order.

\subsection{The Command Manager}

The command manager provides the mechanism by which commands can be undone and redone. It is implemented as a pair of stacks (both initially empty), one of commands that the user has executed/redone and the other of those that have been undone. Its interface (see Table~\ref{tbl:commandmanager}) allows clients to execute commands, undo and redo commands provided the relevant stacks are not empty, check the sizes of the stacks, and clear both of them if desired.

\begin{table}[!p]
\centering
\footnotesize
\ttfamily
\begin{tabular}{|l|l|l|}
\hline
\textbf{CommandManager} & \textbf{consists of} & (executed, undone) \\
\hline
can\_redo() & \textbf{returns} & undone $\ne$ $<>$ \\
\hline
can\_undo() & \textbf{returns} & executed $\ne$ $<>$ \\
\hline
execute\_command(c) & \textbf{ensures} & executed = c : \textbf{old}(executed) \\
                    &                  & undone = $<>$ \\
\hline
executed\_count() & \textbf{returns} & \#executed \\
\hline
redo() & \textbf{requires} & undone $\ne$ $<>$ \\
       & \textbf{ensures} & executed = head(\textbf{old}(undone)) : \textbf{old}(executed) \\
			 &                  & undone = tail(\textbf{old}(undone)) \\
			 & \textbf{invokes} & head(\textbf{old}(undone)).execute() \\
\hline
reset() & \textbf{ensures} & executed = $<>$ \\
        &                  & undone = $<>$ \\
\hline
undo() & \textbf{requires} & executed $\ne$ $<>$ \\
       & \textbf{ensures} & executed = tail(\textbf{old}(executed)) \\
       &                  & undone = head(\textbf{old}(executed)) : \textbf{old}(undone) \\
       & \textbf{invokes} & head(\textbf{old}(executed)).undo() \\
\hline
undone\_count() & \textbf{returns} & \#undone \\
\hline
\end{tabular}
\caption{The \texttt{CommandManager} interface.}
\label{tbl:commandmanager}
\end{table}

Executing a command involves clearing the stack of undone commands, calling the \texttt{execute} member function of the new command and adding it to the stack of executed commands. Undoing a command moves the top element of the executed stack (if any) across to the undone stack and calls its \texttt{undo} member function. Redoing a command moves the top element of the undone stack (if any) across to the executed stack and calls its \texttt{execute} member function. See Figure~\ref{fig:undoredo} for an illustration of the undo/redo process.

\stufig{height=4cm}{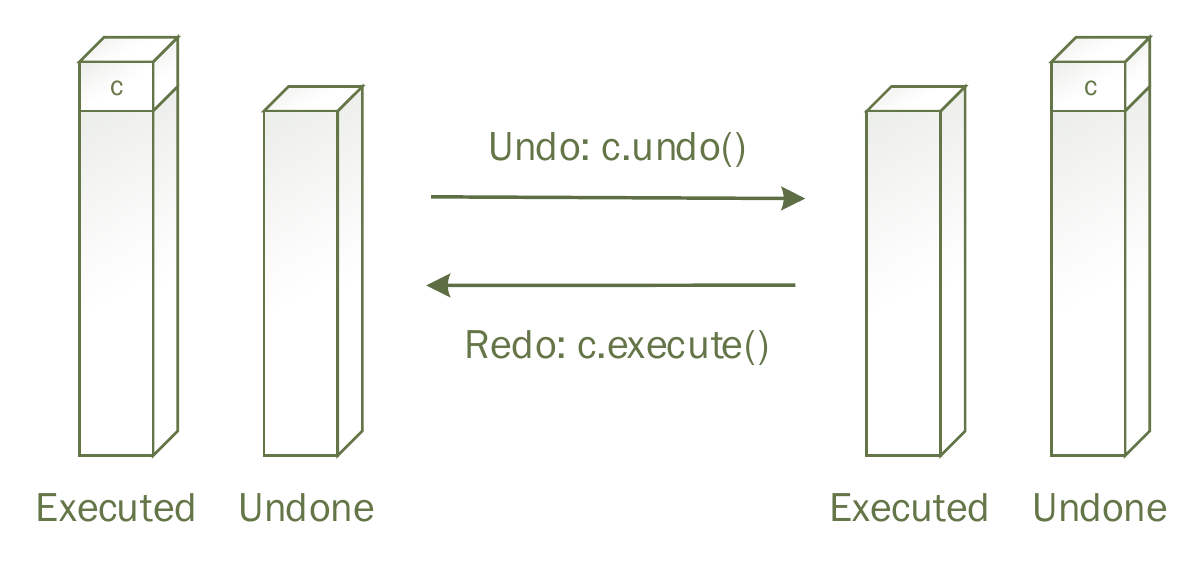}{The undo/redo process in the command manager.}{fig:undoredo}{!p}

\stufig{height=10cm}{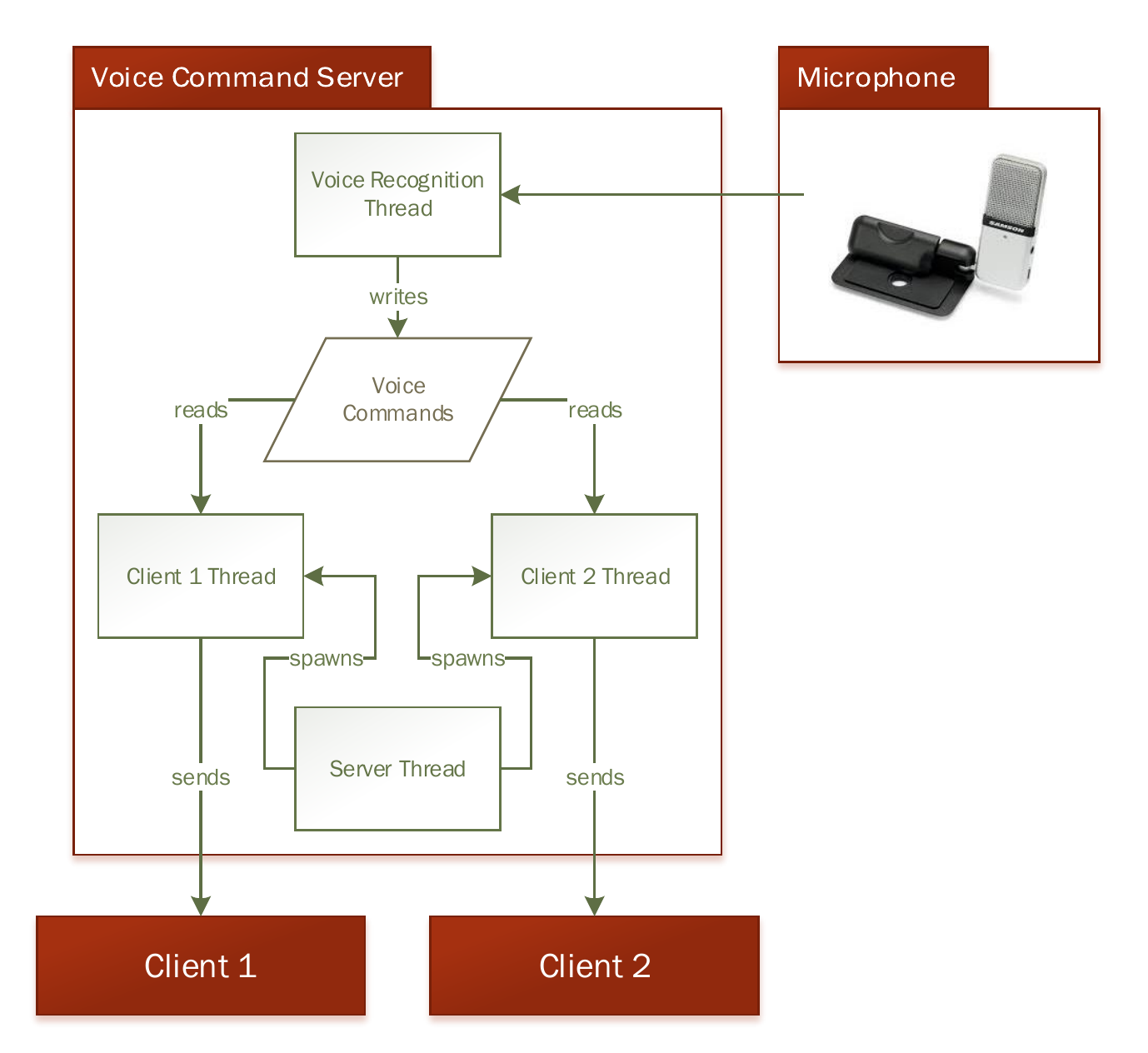}{The architecture of our voice recognition subsystem (slightly simplified to show only $2$ clients).}{fig:voicecommander}{!p}

\section{Voice Recognition}
\label{sec:voicerecognition}

To allow the user to interact verbally with the \emph{SemanticPaint} engine (e.g.~to choose a semantic label with which to label the scene), we implemented a simple, Java-based voice command server called \emph{Voice Commander} using CMU's popular Sphinx-4 speech recognition engine \cite{Lamere2003}.
This listens to voice input from a microphone and attempts to recognise predefined commands spoken by the user.
It accepts connections from any number of clients and notifies them whenever a new command is recognised.
The main \emph{SemanticPaint} engine initially connects to the server, and then checks for new commands once per frame.
Detected commands are used to immediately update the engine's state, e.g.~the command `label chair' causes the semantic label with which the user is labelling the scene to be set to `chair'.
A diagram illustrating the architecture of our voice recognition subsystem is shown in Figure~\ref{fig:voicecommander}.

\section{Touch Detection}
\label{sec:touchdetection}
The detection of touch interactions is a core part of the \emph{SemanticPaint} framework, and is motivated by the potential for humans to teach a machine about various object categories in a natural way, just by touching objects in the scene.
We allow a user to touch objects using their hands, feet or any other object they wish to use for interaction.
The wide range of objects with which the user may interactively label the 3D world stems from a basic assumption that anything moving in front of the static scene is a candidate for touch interaction.

\subsection{Pipeline}

\subsubsection{Overview}

To detect the places in which the user is touching the scene, we look at the differences between the scene at the time of reconstruction, as captured by a depth raycast of our model from the current camera position, and the scene at the current point in time, as captured by the raw depth input from the camera. If we assume that the scenes we are labelling are mostly static, then these differences will tend to consist of a combination of user interactions with the scene and unwanted noise. By postprocessing the differences appropriately, we aim to identify which of them correspond to genuine user interactions with the scene, and label the scene accordingly.

The pipeline we use to detect user interactions is shown in Figure~\ref{fig:touchdetection}. We first construct an image containing the differences between the reconstructed and current scenes, and threshold this to make a binary change mask. We then find the connected components in this mask, and filter them in order to determine the component (if any) that is most likely to represent a user interaction with the scene. Finally, we again look at the differences between the reconstructed and current scenes, and identify the points within this component that correspond to places in which the user is touching the scene. The details of how the various stages of the pipeline work can be found in the following subsections.

\stufig{height=24.5cm}{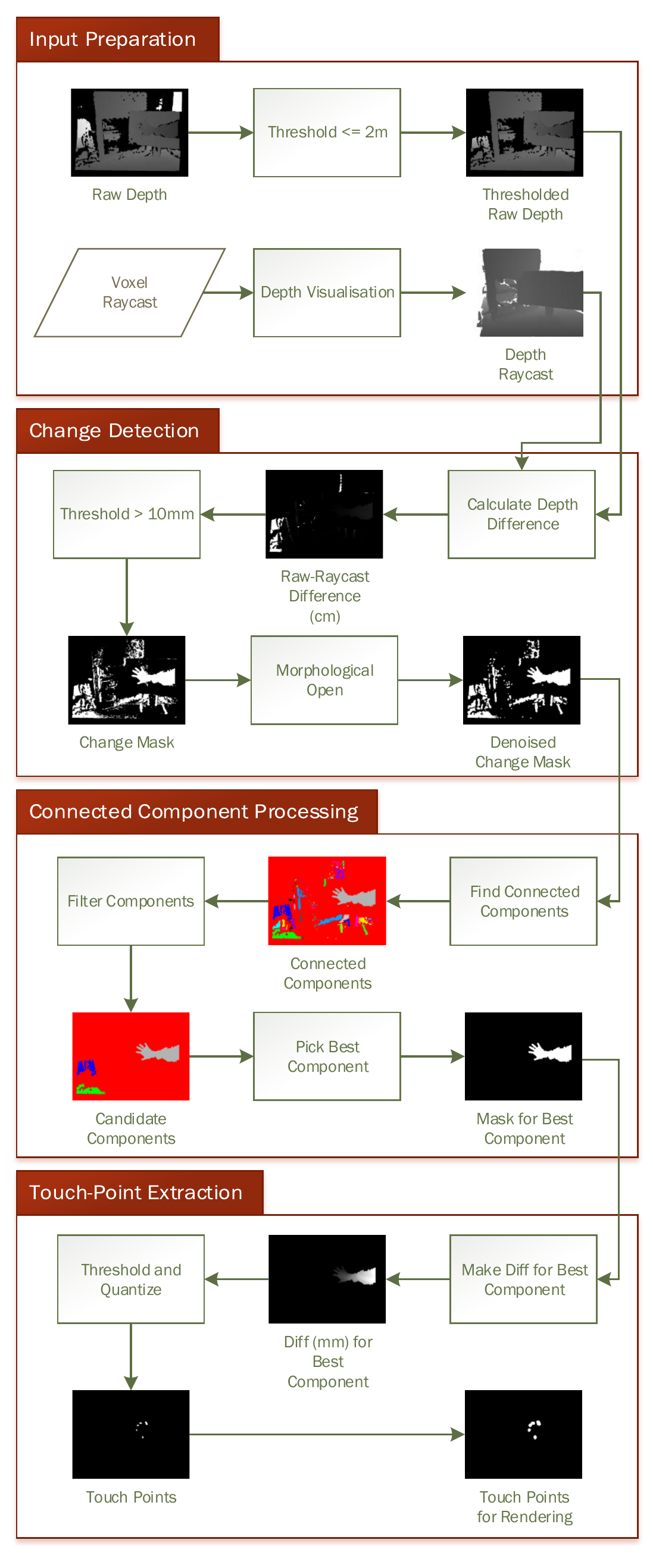}{The touch detection pipeline.}{fig:touchdetection}{!p}

\subsubsection{Input Preparation}
\label{subsec:imagepreprocessing}
\begin{figure}
  \centering

  \subfigure[The thresholded raw depth image.]
  {
    \label{subfig:rawdepth}
    \includegraphics[width=0.45\textwidth]{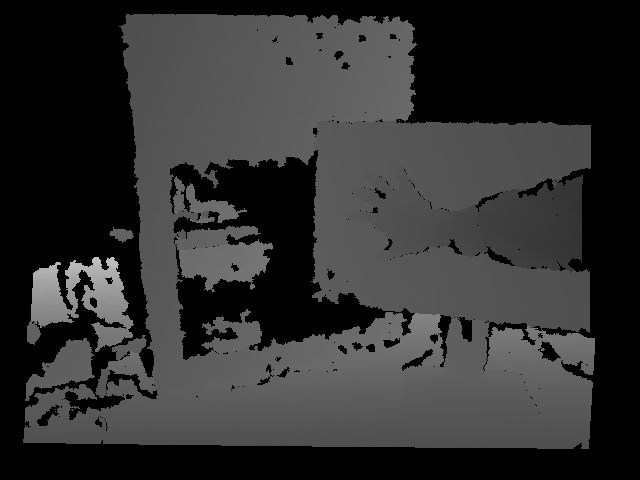}
  }
  \hfill
  \subfigure[A depth raycast of the reconstructed scene.]
  {
    \label{subfig:depthraycast}
    \includegraphics[width=0.45\textwidth]{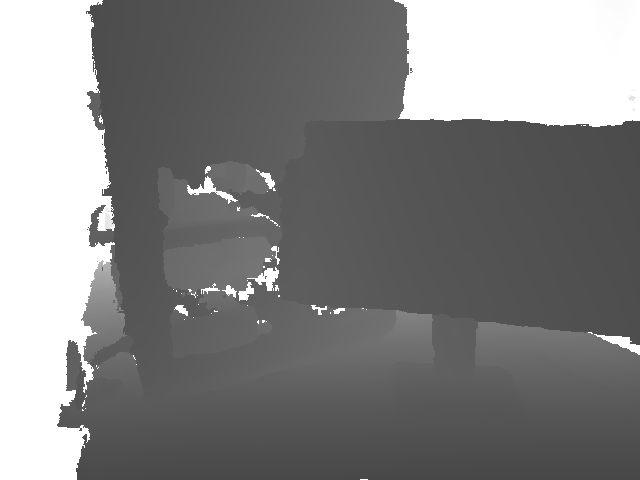}
  }

  \caption{
    Input preparation. \textbf{(a)} The black pixels in the thresholded raw depth image denote regions for which we have no depth, and are set to $-1$.
    \textbf{(b)} The white pixels in the depth raycast of the reconstructed scene denote regions for which the rays we cast into the world did not hit the scene: we assume that this is due to the maximum range limit of the camera, and treat them as being at an arbitrarily large depth (in practice, $100$m).
    The difference between the images in (a) and (b) results in another image that represents the depth offsets of the parts of the scene that have changed with respect to the reconstructed scene.
  }

  \label{fig:pre-processing}
\end{figure}

The inputs to the touch detector are the current raw depth captured by the camera and a voxel raycast of the scene from the current camera position.
The raw depth image is thresholded to ignore parts of the scene that are greater than $2$m away from the camera (see Figure~\ref{subfig:rawdepth}) .
This threshold was chosen to be less than the maximum depth integrated into the scene by InfiniTAM \cite{Kaehler2015},
and to be greater than the maximum expected distance from the camera to a point of interaction (i.e.~we do not expect that the user's hand or foot will extend to more than $2$m away from a camera positioned close to the user).

Based on the voxel raycast, we use orthographic projection to calculate a depth raycast of the reconstructed scene from the current camera position (see Figure~\ref{subfig:depthraycast}).
Pixels for which no valid depth exists are set to an arbitrarily large distance away from the camera (e.g. $100$m).
The alternative approach of ignoring invalid regions would have the undesirable effect of breaking up connected touch components if they occlude parts of the scene that are too far away to be reconstructed.
In the next subsection, the thresholded raw depth and depth raycast will be used to find the parts of the scene that have changed.

\subsubsection{Change Detection}
\label{subsec:changedetection}
The purpose of change detection is to find image regions that have changed since the reconstructed scene was last updated.
Let the thresholded raw depth image be denoted by $\Matrix{D}$ and the depth raycast of the reconstructed scene be denoted by $\Matrix{R}$.
We use these to calculate a binary change mask $\Matrix{C}$ as follows:
\begin{equation}
  \label{eqn:changedetection}
  \Scalar{c}_{\ij} =
	\begin{cases}
	1 & \mbox{if } (\Scalar{d}_{\ij} \ge 0) \land (|\Scalar{d}_{\ij} - \Scalar{r}_{\ij}| > \tau) \\
	0 & \mbox{otherwise}
	\end{cases}
\end{equation}
Here, $\Scalar{m}_{\ij}$ denotes pixel $(\Index{i},\Index{j})$ in image $\Matrix{M}$, and $\tau$ defines a threshold (by default $10$mm) below which the thresholded raw depth and the depth raycast of the reconstructed scene are assumed to be equal.
We test $\Scalar{d}_{\ij}$ against $0$ because a pixel in $\Matrix{D}$ may be $-1$ if the corresponding pixel in the raw depth image was either invalid or greater than the specified $2$m threshold.

After computing the change mask $\Matrix{C}$, we apply a morphological open operation \cite{Gonzalez2006} to it to reduce noise, with the result as shown in Figure~\ref{subfig:changeimage}.
This has the added advantage of reducing the number of connected components that we will calculate in the next subsection,
which in turn reduces the computational cost of subsequent image processing stages.

\subsubsection{Connected Component Processing}
\label{subsec:connectedcomponents}
\begin{figure}
  \centering

  \subfigure[The denoised change mask.]
  {
    \label{subfig:changeimage}
    \includegraphics[width=0.45\textwidth]{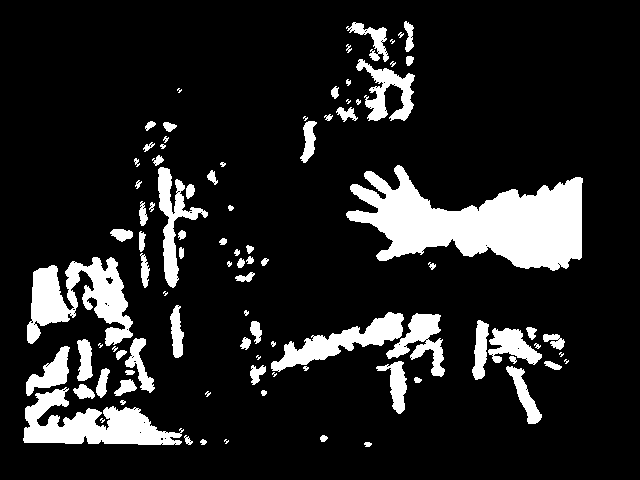}
  }
  \hfill
  \subfigure[The connected component image.]
  {
    \label{subfig:connectedcomponents}
    \includegraphics[width=0.45\textwidth]{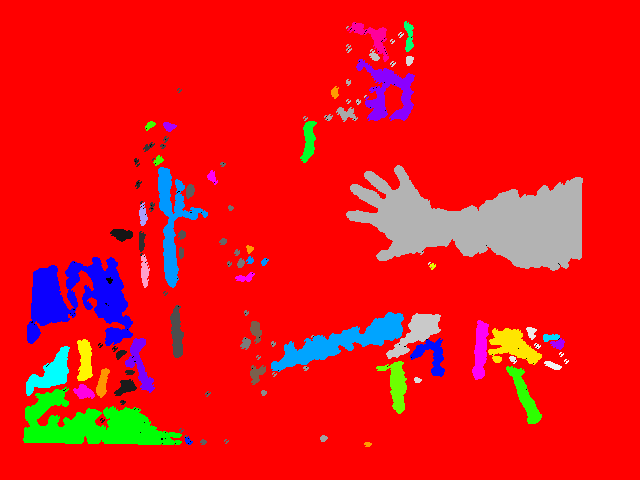}
  }

  \caption{
    \textbf{(a)} The denoised change mask. Note that in spite of the denoising, the change mask still contains significant components that do not correspond to touch interactions, due partly to inaccuracies in the depth measurements, and partly to small errors in the pose of the camera.
    \textbf{(b)} The connected component image calculated from the denoised change mask. Many of the unwanted components in this image can be filtered out by thresholding their areas. The surviving components can be filtered using a random forest trained on unnormalised histograms of connected component difference images.
  }
  \label{fig:connectedcomponents}
\end{figure}

The next step is to calculate (using standard methods \cite{Gonzalez2006}) a connected component image from the denoised change mask.
This is an image in which the pixels in each component share the same integer value (see Figure~\ref{subfig:connectedcomponents}, in which we have applied a colour map to make the different components easier to visualise).

The connected components represent the changing parts of the scene.
They are subsequently filtered to find the component (if any) that is most likely to represent a touch interaction.
The initial filtering stage removes candidate components that are either too large or too small to represent viable interactions.
The probability that each connected component that survives this stage is a touch region is then calculated using a random forest trained on unnormalised histograms of connected component difference images (such as the one shown in Figure~\ref{subfig:bestcandidatediff}) for which the ground truth is known.

Let the trained random forest be represented by a function mapping a feature vector $\Vector{x}_\Index{k}$ for each connected component $\Index{k}$ to the probability
$\Scalar{y}_\Index{k}$ that the connected component region is part of a touch interaction:
\begin{equation}
  \MathFunction{f}(\Vector{x}_\Index{k}) \mapsto \Scalar{y}_\Index{k}
\end{equation}
The connected component $\Index{\hat k}$ most likely to be a touch interaction can then be calculated as follows:
\begin{equation}
  \Index{\hat k} = \underset{\Index{k}}{\mathrm{argmax}} \ \Scalar{y}_\Index{k}
\end{equation}
The most likely connected component is classified as being a touch interaction iff $\Scalar{y}_\Index{\hat k} > 0.5$.
If a touch interaction is detected, the final stage of our pipeline involves finding the points (if any) in that component that are touching a surface, as described in the next subsection.

\subsubsection{Touch-Point Extraction}
\label{subsec:touchpointextraction}

\begin{figure}
  \centering

  \subfigure[The difference image for the touch component.]
  {
    \label{subfig:bestcandidatediff}
    \includegraphics[width=0.45\textwidth]{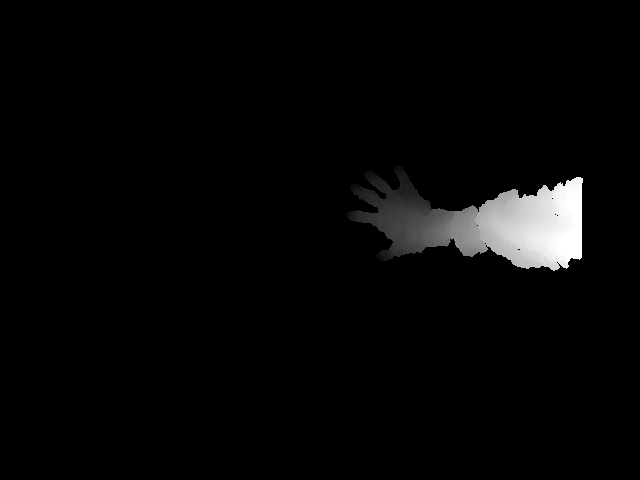}
  }
  \hfill
  \subfigure[The extracted touch points.]
  {
    \label{subfig:touchpoints}
    \includegraphics[width=0.45\textwidth]{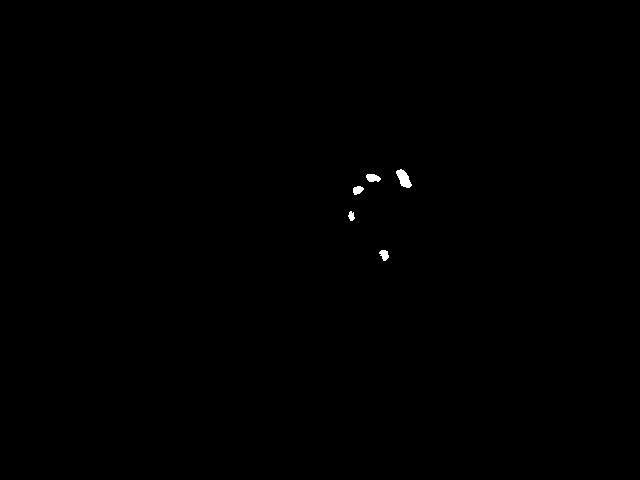}
  }

  \caption{
    The image in (a) shows the depth offset of each pixel in the touch component from the reconstructed scene.
    We threshold this to find the points that are close enough to the surface of the reconstructed scene to be considered as touching it,
    producing the result shown in (b).
  }
  \label{fig:touchpointextraction}
\end{figure}

To extract a set of touch points from a component that has been classified as a touch interaction,
we first construct an image containing the differences between the thresholded raw depth and depth raycast for just those pixels that are within the component (see Figure~\ref{subfig:bestcandidatediff}).
We threshold this image to find points that are within a certain distance of the surface,
producing a result as shown in Figure~\ref{subfig:touchpoints}.

Formally, let the absolute element-wise subtraction between the thresholded raw depth and the depth raycast be an image $\Matrix{S}~=~\left| \Matrix{D} - \Matrix{R} \right|$.
Keeping only the pixels relevant to the selected connected component, we set $\Matrix{S}' = \Matrix{S} \circ \Matrix{C}_{\Index{\hat k}}$,
where $\Matrix{C}_{\Index{\hat k}}$ is a binary mask indicating which elements of $\Matrix{C}$ have value $\Index{\hat k}$, and $\circ$ is the element-wise multiplication operator (i.e.~it applies the binary mask).
The parts of the touch component that are touching the scene may be extracted by setting a threshold $\gamma$ below which it is assumed that the depth difference is small enough to consider that the touch object is in contact with the scene.
The touch point mask is then $\Matrix{T} = \left( \Matrix{S}' > \tau \right) \land \left( \Matrix{S}' < \gamma \right)$,
in which the logical and comparative operators are performed element-wise on matrix $\Matrix{S}'$.
As a final step, since the number of touch points may be quite large, we spatially quantise the image $\Matrix{T}$ by nearest-neighbour down-scaling to extract a smaller set of representative touch points.

\subsection{Discussion}
\label{sec:discussion}
The touch detection algorithm described above is quick enough to run as part of a real-time system, taking approximately $5$ms on each frame, but it has several downsides.
If a hand interacting with the scene hovers over part of the scene that contains inaccurate depth measurements,
those noisy areas may become part of the same connected component as the hand, since no shape constraints are imposed on the interacting object.
Furthermore, the random forest used in the touch detector does not perfectly filter out candidate connected components that are not touch interactions.
We think that the detections may be improved by augmenting the unnormalised histograms of depth difference with more complex connected component features.
Frequent sources of noise arise from inaccurate depth measurements,
caused by objects that absorb/reflect infrared light in ways that disturb the reflected patterns needed by structured light sensors.
Moreover, inserting an object into the scene may interfere with the iterative closest point (ICP) camera pose tracking,
causing unwanted depth differences at the edges of objects.
Finally, the assumption that the moving parts of the scene may be extracted by taking the difference between the thresholded raw depth and the depth raycast rests upon an assumption of highly-accurate camera pose tracking, which does not always hold.
The effects of these nuisance factors limit the reliability of the touch detector in anything other than ideal capturing conditions. In future versions of our framework, we hope to implement alternative approaches to touch detection to account for these difficulties.

\bibliographystyle{alpha}
\bibliography{spainttr}

\newcommand{\etalchar}[1]{$^{#1}$}
\begin{thebibliography}{EGW{\etalchar{+}}10}

\bibitem[Bre01]{Breiman2001}
Leo Breiman.
\newblock {Random Forests}.
\newblock {\em Machine Learning}, 45(1):5--32, 2001.

\bibitem[CCPS15]{Choi2015}
Wongun Choi, Yu-Wei Chao, Caroline Pantofaru, and Silvio Savarese.
\newblock {Indoor Scene Understanding with Geometric and Semantic Contexts}.
\newblock {\em International Journal of Computer Vision (IJCV)},
  112(2):204--220, 2015.

\bibitem[CSK12]{Criminisi2012}
Antonio Criminisi, Jamie Shotton, and Ender Konukoglu.
\newblock {Decision Forests: A Unified Framework for Classification,
  Regression, Density Estimation, Manifold Learning and Semi-Supervised
  Learning}.
\newblock {\em Foundations and Trends{\rm} in Computer Graphics and Vision},
  7(2-3):81--227, 2012.

\bibitem[DZ13]{Dollar2013}
Piotr Doll{\'a}r and C.~Lawrence Zitnick.
\newblock {Structured Forests for Fast Edge Detection}.
\newblock In {\em IEEE International Conference on Computer Vision (ICCV)},
  pages 1841--1848, 2013.

\bibitem[EGW{\etalchar{+}}10]{Everingham2010}
Mark Everingham, Luc~Van Gool, Christopher K~I Williams, John Winn, and Andrew
  Zisserman.
\newblock {The \textsc{Pascal} Visual Object Classes (VOC) Challenge}.
\newblock {\em International Journal of Computer Vision (IJCV)},
  88(2):303--338, 2010.

\bibitem[Gol06]{Golodetz2006}
Stuart~M. Golodetz.
\newblock {A 3D Map Editor}.
\newblock Undergraduate thesis, Oxford University Computing Laboratory, May
  2006.

\bibitem[GSV{\etalchar{+}}15]{Golodetz2015}
Stuart Golodetz, Michael Sapienza, Julien P.~C. Valentin, Vibhav Vineet,
  Ming-Ming Cheng, Victor~A. Prisacariu, Olaf K{\"a}hler, Carl~Yuheng Ren,
  Anurag Arnab, Stephen~L. Hicks, David~W Murray, Shahram Izadi, and Philip
  H.~S. Torr.
\newblock {SemanticPaint: Interactive Segmentation and Learning of 3D Worlds}.
\newblock In {\em ACM SIGGRAPH Emerging Technologies (Demo)}, page~22, August
  2015.

\bibitem[GW06]{Gonzalez2006}
Rafael~C. Gonzalez and Richard~E. Woods.
\newblock {\em Digital Image Processing (3rd Edition)}.
\newblock Prentice-Hall, Inc., Upper Saddle River, NJ, USA, 2006.

\bibitem[HSO07]{Harris2007}
Mark Harris, Shubhabrata Sengupta, and John~D. Owens.
\newblock {Parallel Prefix Sum (Scan) with CUDA}.
\newblock In {\em GPU Gems 3}, chapter~39, pages 851--876. Addison Wesley,
  2007.

\bibitem[KBBP11]{Kontschieder2011}
Peter Kontschieder, Samuel~Rota Bul{\`o}, Horst Bischof, and Marcello Pelillo.
\newblock {Structured Class-Labels in Random Forests for Semantic Image
  Labelling}.
\newblock In {\em IEEE International Conference on Computer Vision (ICCV)},
  pages 2190--2197, 2011.

\bibitem[KPR{\etalchar{+}}15]{Kaehler2015}
Olaf K{\"a}hler*, Victor~Adrian Prisacariu*, Carl~Yuheng Ren, Xin Sun, Philip
  Torr, and David Murray.
\newblock {Very High Frame Rate Volumetric Integration of Depth Images on
  Mobile Devices}.
\newblock {\em IEEE Transactions on Visualization and Computer Graphics
  (TVCG)}, 21(11):1241--1250, 2015.

\bibitem[LFU13]{Lin2013}
Dahua Lin, Sanja Fidler, and Raquel Urtasun.
\newblock {Holistic Scene Understanding for 3D Object Detection with RGBD
  cameras}.
\newblock In {\em IEEE International Conference on Computer Vision (ICCV)},
  pages 1417--1424, 2013.

\bibitem[Lic13]{Lichman2013}
M.~Lichman.
\newblock {UCI Machine Learning Repository}, 2013.

\bibitem[LKG{\etalchar{+}}03]{Lamere2003}
Paul Lamere, Philip Kwok, Evandro Gouv{\^e}a, Bhiksha Raj, Rita Singh, William
  Walker, Manfred Warmuth, and Peter Wolf.
\newblock {The CMU Sphinx-4 Speech Recognition System}.
\newblock In {\em IEEE International Conference on Acoustics, Speech and Signal
  Processing (ICASSP)}, volume~1, pages 2--5, 2003.

\bibitem[NZIS13]{Niessner2013}
Matthias Nie{\ss}ner, Michael Zollh{\"o}fer, Shahram Izadi, and Marc
  Stamminger.
\newblock {Real-time 3D Reconstruction at Scale using Voxel Hashing}.
\newblock {\em ACM Transactions on Graphics (TOG)}, 32(6):169, 2013.

\bibitem[PKC{\etalchar{+}}14]{Prisacariu2014}
Victor~Adrian Prisacariu*, Olaf K{\"a}hler*, Ming~Ming Cheng, Carl~Yuheng Ren,
  Julien Valentin, Philip H.~S. Torr, Ian~D. Reid, and David~W. Murray.
\newblock {A Framework for the Volumetric Integration of Depth Images}.
\newblock {\em ArXiv e-prints}, 2014.

\bibitem[SSK{\etalchar{+}}13]{Shotton2013}
Jamie Shotton, Toby Sharp, Pushmeet Kohli, Sebastian Nowozin, John Winn, and
  Antonio Criminisi.
\newblock {Decision Jungles: Compact and Rich Models for Classification}.
\newblock In {\em Advances in Neural Information Processing Systems (NIPS)},
  pages 234--242, 2013.

\bibitem[Suf04]{Sufrin2004}
Bernard Sufrin.
\newblock {Further Design Patterns (Part II): Active Syntax and Visitor,
  Command}.
\newblock \emph{Object-Oriented Programming} course, Oxford University
  Computing Laboratory, 2004.

\bibitem[Vit85]{Vitter1985}
Jeffrey~Scott Vitter.
\newblock {Random Sampling with a Reservoir}.
\newblock {\em ACM Transactions on Mathematical Software (TOMS)}, 11(1):37--57,
  1985.

\bibitem[VVC{\etalchar{+}}15]{Valentin2015}
Julien Valentin, Vibhav Vineet, Ming-Ming Cheng, David Kim, Shahram Izadi,
  Jamie Shotton, Pushmeet Kohli, Matthias Nie{\ss}ner, Antonio Criminisi, and
  Philip H.~S. Torr.
\newblock {SemanticPaint: Interactive 3D Labeling and Learning at your
  Fingertips}.
\newblock {\em ACM Transactions on Graphics (TOG)}, 34(5), August 2015.

\bibitem[ZZY{\etalchar{+}}15]{Zheng2015}
Bo~Zheng, Yibiao Zhao, Joey Yu, Katsushi Ikeuchi, and Song-Chun Zhu.
\newblock {Scene Understanding by Reasoning Stability and Safety}.
\newblock {\em International Journal of Computer Vision (IJCV)},
  112(2):221--238, 2015.

\end{thebibliography}

\end{document}